\documentclass[sn-basic]{sn-jnl}

\usepackage{geometry}
\usepackage{graphicx}%
\usepackage{multirow}%
\usepackage{amsmath,amssymb,amsfonts}%
\usepackage{amsthm}%
\usepackage{mathrsfs}%
\usepackage[title]{appendix}%
\usepackage{xcolor}%
\usepackage{textcomp}%
\usepackage{manyfoot}%
\usepackage{booktabs}%
\usepackage{algorithm}%
\usepackage{algorithmicx}%
\usepackage{algpseudocode}%
\usepackage{listings}%
\usepackage{algpseudocode,algorithm}
\usepackage{soul}
\usepackage{dsfont}
\usepackage{tabularx}
\usepackage{hyperref}
\usepackage{fancyhdr}

\fancypagestyle{firstpage}{
  \fancyhf{} 
  \rfoot{} 
  \cfoot{} 
  \lfoot{Published in \textit{Machine Learning} by Springer (2023). \href{https://doi.org/10.1007/s10994-023-06454-2}{https://doi.org/10.1007/s10994-023-06454-2}} 
}

\begin{document}

\title[Article Title]{
\textbf{Active learning for data streams: a survey}
}


\author*[1,2]{\fnm{Davide} \sur{Cacciarelli}}\email{dcac@dtu.dk}

\author[1,3]{\fnm{Murat} \sur{Kulahci}}\email{muku@dtu.dk}

\affil[1]{\orgdiv{Department of Applied Mathematics and Computer Science}, \orgname{Technical University of Denmark}, \orgaddress{\city{Kgs. Lyngby}, \country{Denmark}}}

\affil[2]{\orgdiv{Department of Mathematical Sciences}, \orgname{Norwegian University of Science and Technology}, \orgaddress{\city{Trondheim}, \country{Norway}}}

\affil[3]{\orgdiv{Department of Business Administration, Technology and Social Sciences}, \orgname{Luleå University of Technology}, \orgaddress{\city{Luleå}, \country{Sweden}}}


\abstract{Online active learning is a paradigm in machine learning that aims to select the most informative data points to label from a data stream. The problem of minimizing the cost associated with collecting labeled observations has gained a lot of attention in recent years, particularly in real-world applications where data is only available in an unlabeled form. Annotating each observation can be time-consuming and costly, making it difficult to obtain large amounts of labeled data. To overcome this issue, many active learning strategies have been proposed in the last decades, aiming to select the most informative observations for labeling in order to improve the performance of machine learning models. These approaches can be broadly divided into two categories: static pool-based and stream-based active learning. Pool-based active learning involves selecting a subset of observations from a closed pool of unlabeled data, and it has been the focus of many surveys and literature reviews. However, the growing availability of data streams has led to an increase in the number of approaches that focus on online active learning, which involves continuously selecting and labeling observations as they arrive in a stream. This work aims to provide an overview of the most recently proposed approaches for selecting the most informative observations from data streams in real time. We review the various techniques that have been proposed and discuss their strengths and limitations, as well as the challenges and opportunities that exist in this area of research.}




\maketitle
\thispagestyle{firstpage} 

\section{Introduction} \label{sec:introduction}
The deployment of machine learning models in real-world applications is often reliant on the availability of significant amounts of annotated data. While recent advancements in sensor technology have facilitated the collection of larger amounts of data, this data is not always labeled and ready for use in training models. Indeed, the process of obtaining labeled observations for supervised learning models can be cost-prohibitive and time-consuming, as it often requires quality inspections or manual annotation. In such cases, active learning proves to be a valuable strategy to identify the most informative data points for use in training, thereby reducing the overall cost of labeling and improving the performance of the model. Over the years, a plethora of active learning approaches have been proposed in the literature, each with its own benefits and limitations. These approaches seek to strike a balance between the cost of labeling and the quality of the model by selectively choosing the most informative observations for querying. By carefully selecting the most informative observations, active learning helps to minimize the amount of labeled data required and streamlines the learning process, contributing to its overall efficiency. 

While several surveys have been published on pool-based active learning \citep{surveyaggarwal,Settles2009,Fu2013,Kumar2020}, which involves selecting a fixed set of observations from a pool of unlabeled data, the dynamic and sequential nature of many real-world problems often renders these approaches impractical. This has led to growing interest in the online variant of active learning, also referred to as stream-based active learning, which involves continuously selecting and labeling observations as they arrive in a stream, allowing for real-time adaptation to changing data distributions. \cite{Lughofer2017} provided a review of online active learning approaches with a focus on fuzzy models. However, since its publication, numerous other online active learning approaches have been proposed, and to the best of our knowledge, no other surveys have been published to synthesize these developments. Moreover, surveys purely focusing on online learning from data streams \citep{Lu2018,Tieppo2022,Lima2022,Hoi2021} discuss methods that assume a complete availability of labels, which is not the case in many real-world applications. The aim of this review is to fill this gap by providing a comprehensive overview \footnote{We conducted a search on SCOPUS and Google Scholar using the following keywords: "on-line active learning", "online active learning", "stream-based active learning", "single pass active learning", "online selective sampling", "sequential selective sampling", and "active learning" combined with "data stream". Each paper was reviewed individually to determine its relevance to online active learning. We eliminated irrelevant papers and manually added some papers that did not contain these keywords but used online active learning methods or were relevant to our discussion. Additionally, we included related papers that were necessary to understand the bigger picture from the references of the reviewed strategies.} of the most recently developed query strategies for online active learning. It is worth noting that in certain cases, stream-based active learning is narrowly defined as the act of selecting the most informative observations from a data stream to fit a predictive model. Instead, the act of determining which observations to query while making predictions is referred to as online selective sampling \citep{Hanneke2021}. In this work, we cover and examine all the methods that address the crucial problem of selecting the most informative data points to label from a data stream in an online fashion. We will present the techniques that have been proposed so far, discussing their strengths and limitations, as well as the challenges and opportunities that exist in this field. In addition, we will provide an overview of evaluation strategies for online active learning algorithms and highlight some real-world applications. Finally, we will identify potential future research directions in this area. 

This survey comprehensively explores various facets of active learning, encompassing both theoretical foundations and practical challenges. By delving into this review, we aim to shed light on pertinent research questions, including:

\begin{enumerate}
    \item \textit{Query strategy. }What sampling strategy should be used to maximize learning efficiency in a streaming context?
    \item \textit{Timing of queries. }When and how often should data points be queried to balance learning and resource constraints?
    \item \textit{Model updates. }When should predictive models be updated and how can they adapt to changing data distributions and concept drift?
    \item \textit{Scalability. }How can active learning methods be made scalable and efficient for high-velocity data streams?
    \item \textit{Evaluation. }What are appropriate evaluation metrics for assessing the performance of stream-based active learning algorithms?
\end{enumerate}


The structure of this paper is as follows. In Section \ref{sec:preliminaries}, we provide an overview of active learning, including the main instance selection criteria, an overview of the main active learning scenarios, and the connection between active learning and semi-supervised learning. Section \ref{sec:approaches} represents the core of the review, with a brief overview of how online active learning approaches have been classified, followed by a detailed description of the state-of-the-art approaches. In Section \ref{sec:evaluation}, we examine evaluation strategies for online active learning algorithms. Section \ref{sec:application} highlights real-world applications and challenges. Section \ref{sec:summary} provides a summary of the most common online active learning methods and highlights potential directions for future research. Finally, Section \ref{sec:conclusion} provides conclusions and summarizes the key contributions of the review.

\section{Preliminaries on active learning} \label{sec:preliminaries}
In supervised learning, we seek to learn a function that can predict the output variable, also known as response, given a set of input variables, also known as covariates. This function is often learned by training a model on a labeled dataset that consists of a large number of input-output pairs. However, obtaining labeled examples is not always straightforward, and it may not be possible or practical to label all the available data. In these cases, active learning can be used to select a subset of the data for labeling in order to improve the performance of the model, when there is a budget constraint on the number of unlabeled observations that can be queried. Indeed, there are many examples of how a classification or regression model can achieve a performance that is similar to what can be achieved when all the labels are available, using only a small fraction of the available observations.

\subsection{Instance selection criteria}
The main challenge in active learning is deciding which data points to label. There are many strategies for selecting data points in active learning, and most of them can be associated with one of these groups:
\begin{itemize}
    \item \textit{Uncertainty-based query strategies. }These approaches focus on selecting data points that the model is least confident about, in order to reduce its uncertainty \citep{Lu2016,Tong2002}. When using classification models, the most widely used is the margin-based query strategy, where data points close to the decision boundary are selected \citep{Roth2006,Balcan2007}.
    \item \textit{Expected error or variance minimization. }These strategies estimate the future error or variance, when a newly labeled example is made available, and try to minimize it directly \citep{Cohn1996,royerror}. 
    \item \textit{Expected model change maximization. }This strategy involves selecting data points that would have the greatest impact on the estimate of the current model parameters if they were labeled and added to the training set \citep{Cai2013}.
    \item \textit{Disagreement-based query strategies. }These approaches focus on selecting data points where there is disagreement among multiple models or experts \citep{Hanneke2014,Wang2011,Hanneke2014minimax,Sheng2008}. One of the most common approaches that use an ensemble of models is query by committee \citep{Seung1992,Freund1997,Burbidge2007}, which uses an ensemble of models to identify instances where the models have conflicting predictions.
    \item \textit{Diversity- and density-based approaches. }These methods exploit the structural information of the instances and try to select data points that are diverse and representative of the overall distribution of the data. One example of this approach is the use of Mahalanobis distance to seek observations that are far from the currently labeled data points \citep{Ge2014,Cacciarelli2022}. Clustering may be applied to label representative data points \citep{Nguyen2004,Min2020,iencoclustering}, and graph-based methods can be employed to explore the structure information of labeled and unlabeled data points \citep{ZhangGraph} or to build upon the semi-supervised label propagation strategy \citep{Long2008}.
    \item \textit{Hybrid strategies. }These are active learning algorithms that combine multiple instance selection criteria \citep{Donmez2007,Huang2014}. For example, by combining margin-based sampling with clustering the learner can select the most uncertain observations within different areas of the input space.
\end{itemize}

By considering these different strategies, one can select the most appropriate approach for a given problem based on the characteristics of the data and the specific requirements of the application.

\subsection{Active learning scenarios}
Active learning can be broadly categorized into three macro scenarios, based on how the unlabeled instances are supplied to the learner and then selected to be labeled by an oracle. Regardless of the particular query strategy being employed, these macro scenarios provide a framework for understanding the flow of information and the decision-making steps involved in active learning. These scenarios serve as a high-level categorization of different methods for approaching the active learning problem, each with its own set of advantages and disadvantages depending on the specific use case. Understanding these macro scenarios is crucial for selecting the appropriate active learning technique for a particular problem and for comparing different active learning algorithms. In the next subsections, each of the three macro scenarios will be discussed.

\subsubsection{Membership query synthesis active learning}
\begin{figure}[h]
  \centering
  \includegraphics[width=0.7\linewidth]{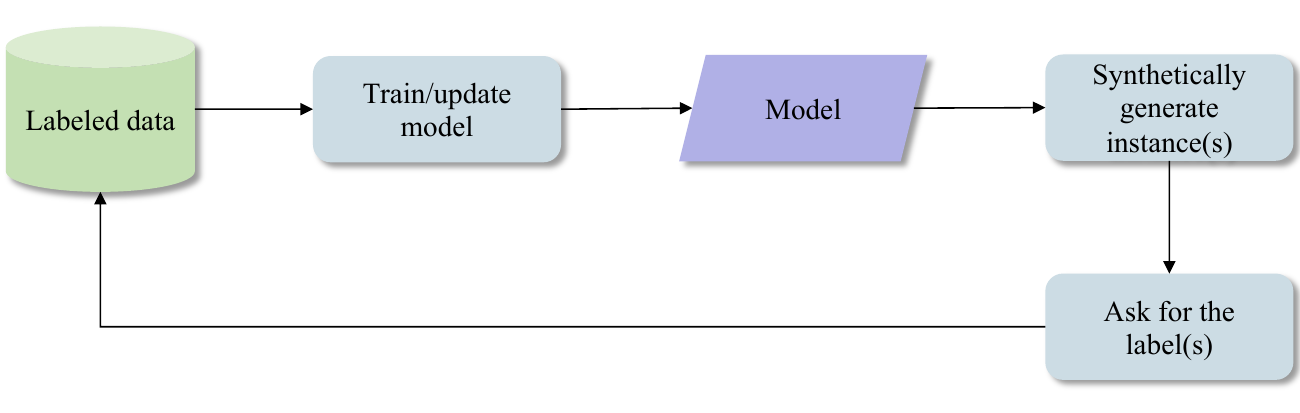}
  \caption{Membership query synthesis active learning.}
  \label{fig:membership}
\end{figure}

This scenario represents the case when the learner is given complete freedom to ask for the label of any data point belonging to the input space or for a synthetically generated one. Some examples of membership query synthesis active learning include image classification, where the learner can generate modified versions of existing images to be labeled, or object detection, where the learner can generate new instances by combining and transforming existing instances. In natural language processing (NLP) tasks such as text classification or sentiment analysis, the learner might generate synthetic examples in the form of sentences or paragraphs that cover a wider range of variations in the language. Also, in speech recognition, the learner might generate synthetic speech samples in different accents, pronunciations, or speaking styles in order to improve the recognition accuracy. However, as highlighted by \cite{Baum1992} and \cite{Settles2009}, the main drawback of this strategy is that it could generate unlabeled examples for which no labels can be associated by a human annotator (e.g., a mixture between a number and a letter). A general flowchart for this scenario is reported in Figure \ref{fig:membership}, where the scheme is repeated until a budget constraint on the requested labels is met, or a stopping criterion on the achieved performance is satisfied.

In the context of deep active learning \citep{Ren2022}, the membership query synthesis scenario can be addressed by using generative models. For instance, generative adversarial networks (GANs) have been used to generate additional instances from the input space that may provide more informative labels for the learner \citep{Goodfellow2014}. This can be done by using GANs for data augmentation, as GANs are capable of generating diverse and high-quality instances \citep{Zhu2017}. Another approach is to combine the use of variational autoencoders (VAEs) \citep{Kingma2013} and Bayesian data augmentation, as demonstrated by Tran et al. \citep{Tran2019, Tran2017}. The authors used VAEs to generate instances from the disagreement regions between multiple models, and Bayesian data augmentation to incorporate the uncertainty of the generated instances in the learning process.

\subsubsection{Pool-based active learning}
Pool-based active learning is one of the most widely studied scenarios in the machine learning literature. The goal is to select the most informative subset of observations from a closed, static set of unlabeled data points. The majority of the proposed pool-based active learning approaches have been developed for classification tasks \citep{Cai2013}, with image classification being a common application in computer vision \citep{Li2013}, as manually labeling large image datasets can be a challenging task.

\begin{figure}[h]
  \centering
  \includegraphics[width=0.7\linewidth]{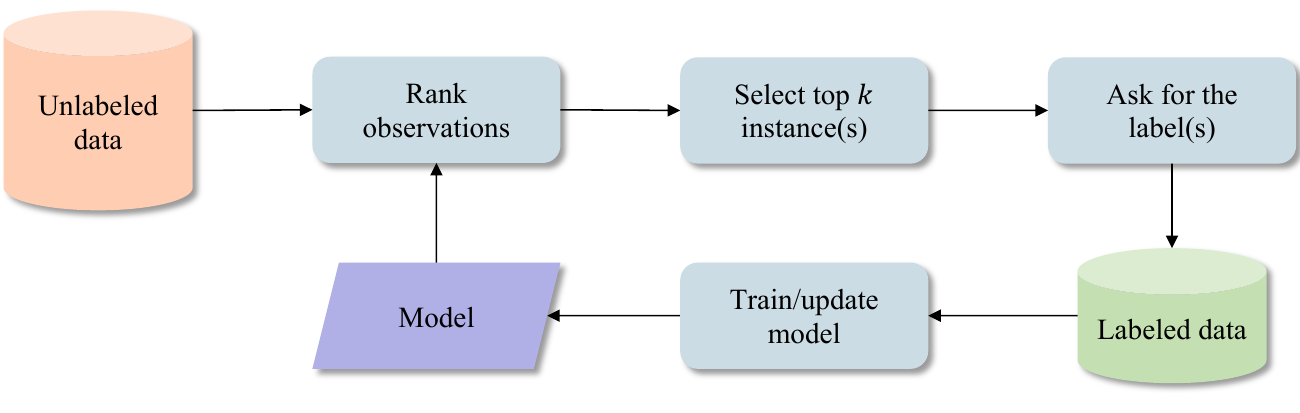}
  \caption{Pool-based active learning.}
  \label{fig:pool_based}
\end{figure}

The flowchart in Figure \ref{fig:pool_based} provides an overview of pool-based active learning sampling schemes, where $k$ represents the number of unlabeled instances whose label is queried at each round. Traditional machine learning models that do not require substantial computational resources to train are typically associated with a choice of $k$ equal to one \citep{Vahdat2019}. This allows a timely update of the instance selection criteria, avoiding the redundant labeling of similar data points. However, larger values of $k$ have also been used in practice, such as the analysis performed by \cite{Ge2014} for values ranging from 5 to 30 or the approach used by \cite{Cai2013} to add 3\% of the total number of observations to the training set each time. Using a higher $k$ value may be more practical when working with large models, as repeated training can be computationally expensive and challenging. To this extent, batch mode active learning is generally considered to be a more efficient and effective option for image classification or detection tasks compared to the one-by-one query strategy, as the latter can be resource-intensive and time-consuming when working with large neural networks \citep{Ren2022}. This is because re-training the model with just one new data point with high input dimensionality may not result in significant improvement \citep{Ren2022}. In general, the choice of $k$ may be problem- or model-specific, as it represents a trade-off between computational efficiency and the risk of querying redundant labels. 

To enhance pool-based active learning, many approaches combine uncertainty-based instance selection criteria with acquisition functions such as entropy \citep{Shannon1948,Wu2022}, mutual information \citep{Haussmann2020}, or variation ratio \citep{Schmidt2020}. Entropy is commonly used as an acquisition function in active learning because it provides a way to measure the uncertainty of the model predictions for a given data point. The entropy of a probability distribution is a measure of the amount of disorder or randomness in the distribution. In the context of active learning, the entropy of a model's predicted class probabilities for a data point can be used as a measure of the model's uncertainty about the correct class label for that data point. Acquiring examples with the highest uncertainty is one way to select data points for annotation, but it is not the only way. Mutual information and variation ratio can also be used on the predictions obtained with the current model, in order to seek a diverse set of data points for which the predictions are the most uncertain. For a more comprehensive discussion on pool-based active learning, readers are referred to the surveys \citep{surveyaggarwal,Settles2009,Fu2013,Kumar2020}.


\subsubsection{Online active learning} \label{subsubsec:oal}
In this type of active learning, we cannot greedily select the most informative observations from a static pool, as the instances are generated in a continuous stream and cannot be stored in their entirety before a decision is made. This is similar to the famous statistical puzzle known as the secretary problem \citep{Freeman1983}, where a hiring manager must make a hiring decision for each applicant as they are interviewed, without the benefit of seeing all applicants first. In general, online active learning is a crucial scenario for various real-world applications where the ability to make a sampling decision in real-time is of utmost importance. A few examples are:
\begin{itemize}
    \item \textit{Chemical or manufacturing processes. }In these applications, a learner is tasked with predicting the quality of the final product but may only have a short timeframe to make the sampling decision, to avoid traceability issues, particularly in high-volume production \citep{schmitt2013traceable,lieber2012sustainable}. Also, tasks like predictive maintenance and visual inspection might benefit from a real-time selection of new examples to be labeled and included in the training set \citep{rozanec}.
    \item \textit{Video streaming and clinical trials. }In these cases, a decision must be made on the fly, as users arrive or volunteers appear sequentially, and there may not be enough time to accumulate a pool of potential users or patients \citep{fowler2023clinical,Riquelme2017}.
    \item \textit{Text classification: }In NLP, online active learning can be used for tasks such as sentiment analysis and spam detection, where the learner continuously learns from new incoming data points which need to be labeled to update the model in real-time and improve accuracy \citep{kranjc2015active}.
    \item \textit{Fraud detection. }To effectively detect fraudulent activities, the learner must continuously select new examples to label so that it can continuously update its decision-making process \citep{carcillo2018streaming,carcillo2017assessment}.
    \item \textit{Online customer service. }Online customer service agents can use online active learning to improve their performance by continuously learning from customer interactions. To do this, the learner must continuously select new examples to label or customer information to obtain, so that it can predict the best response based on past interactions and improve its accuracy over time \citep{zheng2006selectively}.
    \item \textit{Marketing. }Online active learning could also be applied in the field of marketing to select informative examples in real-time and continuously optimize customer targeting and personalization \citep{carnein2019customer,jamil2016churn}.
\end{itemize}

\begin{figure}[h]
  \centering
  \includegraphics[width=0.7\linewidth]{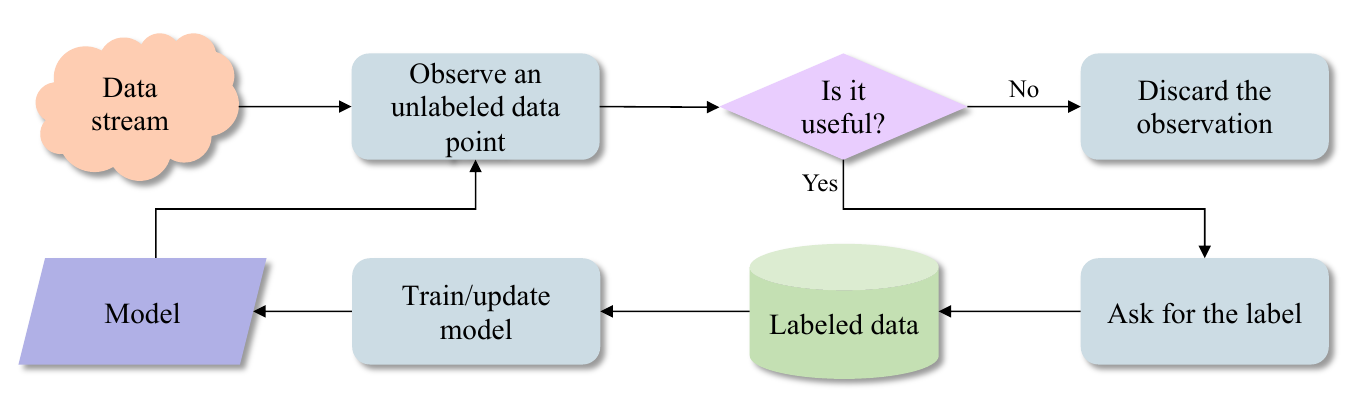}
  \caption{Single-pass online active learning.}
  \label{fig:single_pass}
\end{figure}

\begin{figure}[h]
  \centering
  \includegraphics[width=0.7\linewidth]{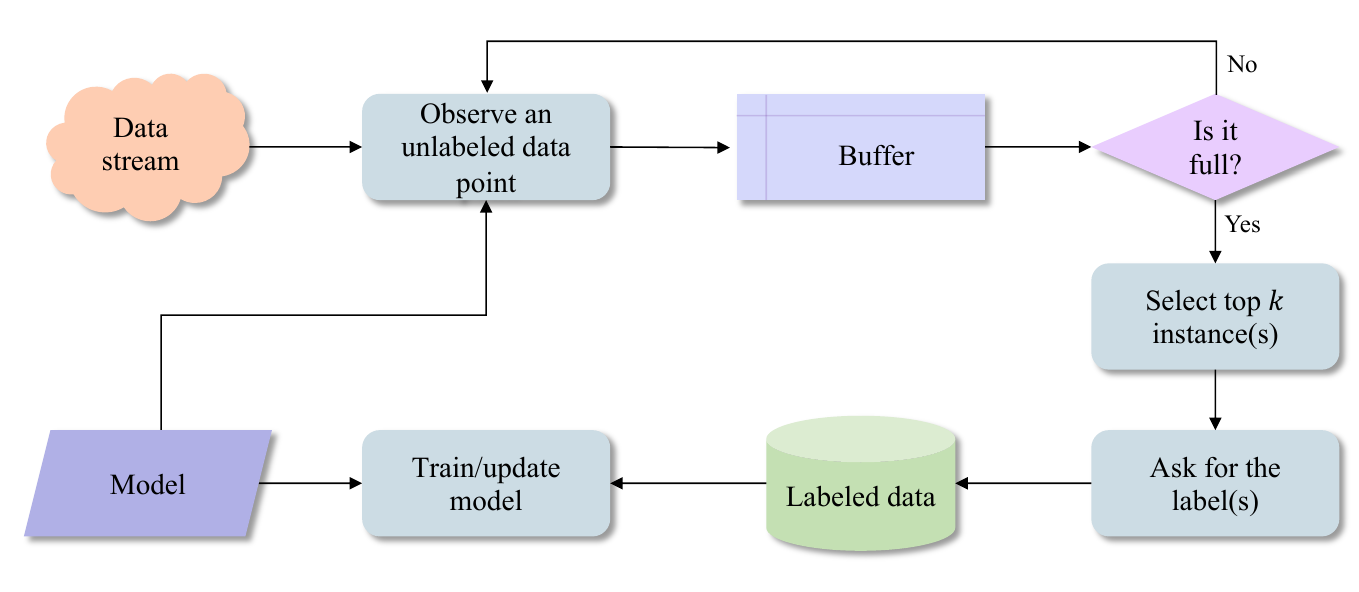}
  \caption{Window- or batch-based online active learning.}
  \label{fig:window_based}
\end{figure}

One of the defining features of online active learning strategies is their data processing capabilities. Figure \ref{fig:single_pass} and Figure \ref{fig:window_based} provide a visual representation of the two main approaches; single-pass and window-based. Single-pass algorithms observe and evaluate each incoming data point on the fly, whereas window-based algorithms, also referred to as batch-based methods, observe a fixed-size chunk of data at a time. In this approach, the learner evaluates the entire batch of data and selects the top $k$ observations as the most informative ones to be labeled. This approach is referred to as best-out-of-window sampling. The specific value of $k$ and the dimensionality of the buffer can vary based on the storage capabilities of the system and the computational time required to update the model. Window-based methods are useful in situations where data is generated in large quantities and the algorithm does not have a tight constraint on the time available for decision-making. In contrast, single-pass methods are necessary when the algorithm needs to make a decision immediately after observing a specific data point. 

Another critical property in the design of an effective online active learning strategy is the assumption made about the data stream distribution. One important difference to consider is whether the data stream is stationary or drifting. A stationary data stream is characterized by a stable data generating process where the statistical properties of the data distribution that remain constant over time. Conversely, a drifting data stream is marked by changing statistical properties of the data distribution over time, potentially due to alterations in the underlying data generating process. The distinction between stationary and drifting data streams is significant because it affects the performance of the active learning strategies. Online active learning strategies that have been developed for stationary data streams may lead to suboptimal performance when applied to drifting data streams. This is because concept drift can alter the scale of the informativeness measure of unlabeled data points or even urge a complete change of the model, with the acquisition of more observations to accommodate the new concept. Therefore, it is important to accurately assess the nature of the data stream distribution in the design of an active learning strategy. A failure to do so can result in a suboptimal performance and a reduced ability to effectively leverage the strengths of active learning. Another important property to consider when designing an active learning strategy is the label delay or verification latency. This refers to the time needed by the oracle to provide the label when it is requested by the learner. In some cases, there may be a delay $L$ in the oracle providing the label after it has been requested. This property must be taken into account when designing a sampling strategy as there may be redundant label requests for similar instances if this issue is not properly addressed. Label delay can be classified into null latency, intermediate latency, or extreme latency \citep{Souza2018}. The case with null latency, or immediate availability of the label upon request, is commonly used in the stream mining community, but may not be realistic for many practical applications. Extreme latency, where labels are never made available to the learner, is closer to an unsupervised learning task. Intermediate latency assumes a delay $0 < L < \infty$ in the availability of the labels from the oracle.

Finally, the training efficiency of the online active learning algorithms should also be taken into consideration. There are two main training approaches in active learning; incremental training and complete re-training. Incremental training involves updating model parameters with a small batch of new data, without starting the training process from scratch \citep{Polikar2001,Wu2019,Shilton2005,Istrate2018}. This approach allows the model to learn from new data while preserving its existing knowledge. This can be achieved through fine-tuning the model parameters with the new data, or by using techniques such as elastic weight consolidation, which prevent previous knowledge from being erased. Complete re-training, on the other hand, involves training a new model from scratch using the entire labeled data collected so far. This approach discards the previous knowledge of the model and starts anew, which may result in the loss of knowledge learned from previous data. Complete re-training is typically used when the amount of new data is substantial, the previous model is no longer relevant, or when the model architecture needs to be altered. It is important to note that the choice of training approach in online active learning algorithms can have a significant impact on the overall performance and effectiveness of the model. 

\subsection{Connection between active learning and semi-supervised learning} \label{subsec:ssl}
Semi-supervised learning is a field of research that is closely related to active learning, as both methods are developed to deal with limited labeled data. While active learning aims to minimize the amount of labeled data required to train a model, semi-supervised learning is a technique that trains a model using a combination of labeled and unlabeled data. Active learning can be considered a special case of semi-supervised learning, as it allows the model to actively select which data points it wants to be labeled, rather than relying on a fixed set of labeled data. In the context of online learning, \cite{Kulkarni2016} conducted a study that provided an overview of semi-supervised learning techniques for classifying data streams. These techniques do not address the primary question of active learning, which is \textit{when to query}, but they are useful in exploiting the information contained in the unlabeled data points and in addressing issues related to model update and retraining in limited labeled data environments. It is also worth noting that semi-supervised learning can be used in combination with active learning to improve the data selection strategy. By leveraging the strengths of both methods, it is possible to achieve better performance and more efficient learning compared to using either method alone.

Semi-supervised learning approaches can be distinguished into three categories, unsupervised preprocessing, wrapper methods, and graph-based methods. Unsupervised preprocessing refers to the use of unsupervised learning techniques, such as dimensionality reduction \citep{cacciarelli2023hidden}, clustering, or feature extraction, to preprocess the entire dataset, labeled and unlabeled, before it is fed to the supervised model \citep{Frumosu2018}. The goal is to transform the data into a more useful representation that can be learned more easily by a supervised model and can support the sampling of more informative data points. This strategy can also help reduce the dimensionality of the learning problem, thus improving the model parameter estimation when only a few queries can be made. Related to the online active learning problem, \cite{rozanec} used a pre-trained network to extract salient features from unlabeled images before starting the sampling routine. Similarly, \cite{Cacciarelli2022} used an autoencoder trained on all the available unlabeled data points to improve the performance of online active learning for linear regression models. 

Wrapper methods, on the other hand, use one or more supervised learners that are trained on labeled data and pseudo-labeled unlabeled data. There are two main variants of wrapper methods, self-training and co-training. Self-training uses a single supervised model that is trained on labeled data, and pseudo-labels are used for the data points with confident predictions. Co-training, on the other hand, extends self-training to multiple supervised models, where two or more models exchange the most confident predictions to obtain pseudo-labels. Pseudo-labels can be very beneficial in label-scarce environments, but one must be mindful of the confirmation bias issue, where the model might rely on incorrect self-created labels. This problem has been extensively analyzed by \cite{Baykal2022} in the active distillation scenario, which is a strategy where a smaller model, known as the student model, is trained to mimic the behavior of a larger pre-trained model, known as the teacher model \citep{Hoang2021,Kwak2022}. In this context, confirmatory bias refers to the student model tendency to reproduce the predictions of the teacher model, even when the teacher predictions are incorrect. This can happen when the student model is trained to mimic the teacher model output too closely, without considering the underlying errors. To mitigate this, active distillation techniques use sample selection methods that encourage the student model to learn from data points where the teacher model makes errors, rather than just reproducing the teacher model predictions. In the more general active learning framework, confirmation bias might also refer to the tendency of an active learning algorithm to select examples that confirm its current hypothesis, rather than selecting examples that would challenge or improve it. 

Finally, graph-based methods construct a graph on all available data and fit a supervised model, where the loss comprises a supervised loss and a regularization term that penalizes the difference between the labels predicted for connected data points. In the online active learning scenario, the graph structure can be used to model the similarity between data points, and the active learning algorithm can select the examples to label based on their position on the graph, such as selecting examples that are in low-density regions or are distant from other labeled examples.


\section{Online active learning approaches} \label{sec:approaches}
In this review, we present a taxonomy of online active learning strategies into four categories:
\begin{enumerate}
    \item \textit{Stationary data stream classification approaches. }These methods are designed to tackle online classification tasks, where the model is updated on the fly using newly labeled examples selected from a stream of data that does not change significantly over time. These methods are particularly useful in scenarios where the data distribution is relatively stable, such as quality control in industrial processes, where stationarity is often ensured by control actions taken at regular intervals and continuous maintenance of the components of the system \citep{bisgaard2011time}. Another example is represented by human activity recognition using wearable devices, where data is collected over time from wearable devices such as fitness trackers to identify patterns of activity like walking, running, or sleeping. This scenario would fall into this category because the data stream is relatively stable, and the model can be updated in real-time as new labeled examples become available \citep{miu2015bootstrapping}.
    \item \textit{Drifting data stream classification approaches. }These online active learning strategies are specifically designed to handle classification tasks in dynamic environments where the data distribution constantly changes. These approaches are designed to adapt to changes in the data distribution in order to maintain high classification accuracy. Some real-world applications might be fraud detection or intrusion detection. In financial fraud detection, fraudsters often change their methods to evade detection, so a classification model used for fraud detection must be able to adapt to new patterns of fraud as they emerge or to new customer habits \citep{Zhang2022}. In real-time intrusion detection, computer networks detection systems must be able to detect new forms of cyberattacks as they appear, so the classification models used must be able to adapt to changes in the data distribution over time \citep{nixon2021reviews}. This scenario would fall into this category because the data stream is constantly changing, and the model must be able to adapt to changes in the data distribution over time to maintain high accuracy.
    \item \textit{Evolving fuzzy system approaches. }These approaches are based on a type of fuzzy system that can adapt and change over time, in response to new data or changes in the environment \citep{gu2023autonomous}. In traditional fuzzy systems, the rules and membership functions that define the system are fixed and do not change over time. Evolving fuzzy systems, on the other hand, are able to adapt their rules and membership functions based on new data or changes in the environment. This is particularly useful in applications where the data or the environment is non-stationary and evolves over time, such as in control systems for autonomous vehicles, where we must be able to adapt to changes in the environment, such as traffic patterns, road conditions, and weather \citep{naranjo2007using,wang2015lateral}.
    \item \textit{Experimental design and bandit approaches. }These methods, mostly related to regression models, actively select the most informative data points to improve model predictions. This category includes online active linear regression and sequential decision-making strategies like bandit algorithms or reinforcement learning. These methods adaptively select the most promising options in a given situation. An example is given by online advertising, where a model is used to select the most promising advertisements to display to users based on their browsing history and other factors \citep{avadhanula2021stochastic}. This scenario would fall into this category because the model must adaptively select the most promising options in real-time based on the information available at that time. Also, in clinical trials, a model is used to select the most promising patients to enroll in a clinical trial based on their medical history and other personal information. Finally, in drug development studies \citep{reda2020machine}, online active learning can be used to select the most promising compounds for further testing and development, based on their potential efficacy and safety. 
\end{enumerate}
This categorization provides a comprehensive overview of the different types of online active learning strategies and how they can be applied in various scenarios. While the simplest active learning strategy, random sampling, is available and involves selecting data points randomly from the stream for annotation, we will primarily focus on more specialized strategies designed to address scenarios where informed decisions are crucial due to resource constraints or where the data distribution is non-stationary.

\begin{figure}[h]
  \centering
  \includegraphics[width=0.8\linewidth]{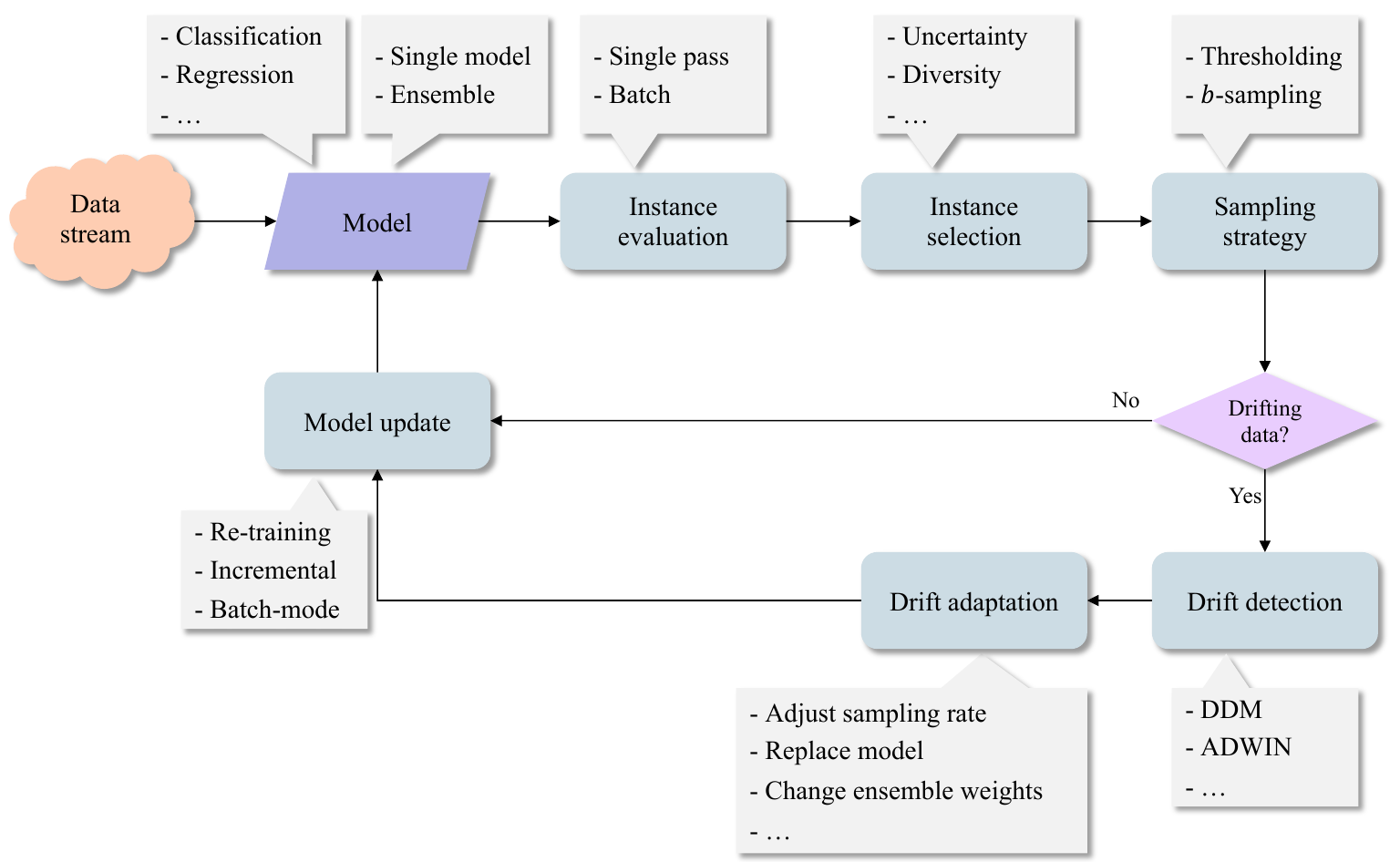}
  \caption{Online active learning: general framework.}
  \label{fig:framework}
\end{figure}

Figure \ref{fig:framework} depicts a general framework illustrating the essential components shared by the various categories of online active learning algorithms. The accompanying callouts highlight key options utilized by these methods. The following sections will provide an in-depth analysis of these strategies. For a more detailed flowchart regarding the drift detection and adaptation process, please refer to \cite{Lu2018,Lima2022}.

\subsection{Stationary data stream classification approaches}
In online active learning, a commonly employed strategy is to request labels for data points that are considered to be informative enough based on a pre-determined threshold. This threshold can be established through a variety of techniques, depending on the instance selection criterion used to evaluate the informativeness of the unlabeled observations. Another method, sometimes referred to as $b$-sampling, is to calculate the probability that a data point will be queried by adjusting the parameter of a Bernoulli random variable, as proposed by Cesa-Bianchi et al. in one of the pioneering studies on online active learning \citep{cesabianchi1,cesabianchi2}. They used a linear predictor characterized by the weight vector $ \mathbf{w} \in \mathds{R}^d$ and, at each time step $t$, after observing the current data point $\mathbf{x}_t$, the binary output $y \in \{-1,+1\}$ is predicted using

\begin{equation}
  \widehat{y}_t=\operatorname{SGN}\left(\mathbf{w}_{t-1}^\top \mathbf{x}_t\right)
\end{equation}

\noindent
where $\mathbf{w}_{t-1}$ is the weight vector estimated with the previously seen labeled examples $ \left(\mathbf{x}_1, y_1\right), \allowbreak \ldots,\allowbreak \left(\mathbf{x}_{t-1}, y_{t-1}\right)$. The value $\mathbf{w}_{t-1}^\top \mathbf{x}_t$ is the margin, $\widehat{p}_t$, of $\mathbf{w}_{t-1}$ on the instance $\mathbf{x}_t$. If the learner queries the label $y_t$, a new weight vector is estimated using the newly added labeled example $\left(\mathbf{x}_t, y_t\right)$ with the regular perceptron update rule \citep{Rosenblatt1958} as in

\begin{equation}
    \mathbf{w}_t=\mathbf{w}_{t-1}+M_t y_t \mathbf{x}_t
\label{eq:perceptron}
\end{equation}

\noindent
where $M_t$ represents the indicator function of the event $\widehat{y}_t \neq y_t$. If the label is not requested, the model remains unchanged, and we have $\mathbf{w}_t=\mathbf{w}_{t-1}$. At each time step $t$, the learner decides whether to query the label of a data point $\mathbf{x}_t$ by drawing a Bernoulli random variable $Z_t \in\{0,1\}$, whose parameter is given by 

\begin{equation}
    P_t=\frac{b}{b+\left|\widehat{p}_t\right|}
\label{eq:bernoulli}
\end{equation}

\noindent
where $b>0$ is a positive smoothing constant that can be tuned to adjust the labeling rate. In general, as $\widehat{p}_t$ approaches $0$, the sampling probability $P_t$ converges to $1$, suggesting that the labels are requested for highly uncertain observations. The sampling scheme introduced by \cite{cesabianchi1} is referred to as selective sampling perceptron, and it is reported in Algorithm \ref{alg:1}.

\begin{algorithm}[h]
\caption{Selective sampling perceptron}\label{alg:1}
\begin{algorithmic}[h]
\Require a data stream $\mathbf{S}$, an initial model $\mathbf{w}_0=(0, \ldots, 0)^\top$, a time horizon $T$, a sampling budget $B$, a parameter $b$.
\State $t \gets 1$ \Comment{Timestamp}
\State $c \gets 0$ \Comment{Labeling cost}
\While{$c \leq B$, $t \leq T$}
    \State Observe an incoming data point $\mathbf{x}_t \in \mathbf{S}$ and set $\widehat{p}_t=\mathbf{w}_{t-1}^\top \mathbf{x}_t$
    \State Predict the label $\widehat{y}_t=\operatorname{SGN}\left(\widehat{p}_t\right)$
    \State Draw a Bernoulli random variable $Z_t$ of parameter $P_t=b /\left(b+\left|\widehat{p}_t\right|\right)$
    \If {$Z_t = 1$} \Comment{Sampling decision}
        \State Ask for the true label $y_t$ and update the model
        \State $c \gets c + 1$ \Comment{Pay for the label}
    \Else
        \State Discard $\mathbf{x}_t$
    \EndIf
    \State $t \gets t + 1$
\EndWhile
\end{algorithmic}
\end{algorithm}

A similar approach to the one proposed by \cite{cesabianchi1} was investigated by \cite{Dasgupta2005}, who presented one of the first thresholding techniques for online active learning. They suggested setting a threshold on the margin, with the idea of sampling data points $\mathbf{x}_t$ with a value of $\left|\widehat{p}_t\right|$ lower than a given threshold $\Gamma$. The threshold is initially set at a high value and iteratively divided by two until enough misclassifications occur among the queried points. The linear classifier is updated using the reflection concept [60] to give more focus to recent data points. \cite{Sculley2007} built on the works of Cesa-Bianchi and Dasgupta to analyze the online active learning scenarios for real-time spam filtering. The author compares two models, a perceptron and a support vector machine (SVM), and tries three different instance selection criteria, the fixed thresholding approach by \cite{Dasgupta2005}, the Bernoulli-based approach by \cite{cesabianchi1}, and a newly developed logistic margin sampling. The perceptron is updated as per \cite{Dasgupta2005}, while the SVM is retrained on all available labeled observations each time a new data point is added. According to the logistic margin sampling strategy, the sampling decision is taken by drawing a Bernoulli random variable $Z_t \in\{0,1\}$ with a parameter given by

\begin{equation}
    P_t=e^{-\gamma\left|\widehat{p}_t\right|}
\end{equation}

\begin{figure}[h]
  \centering
  \includegraphics[width=.8\linewidth]{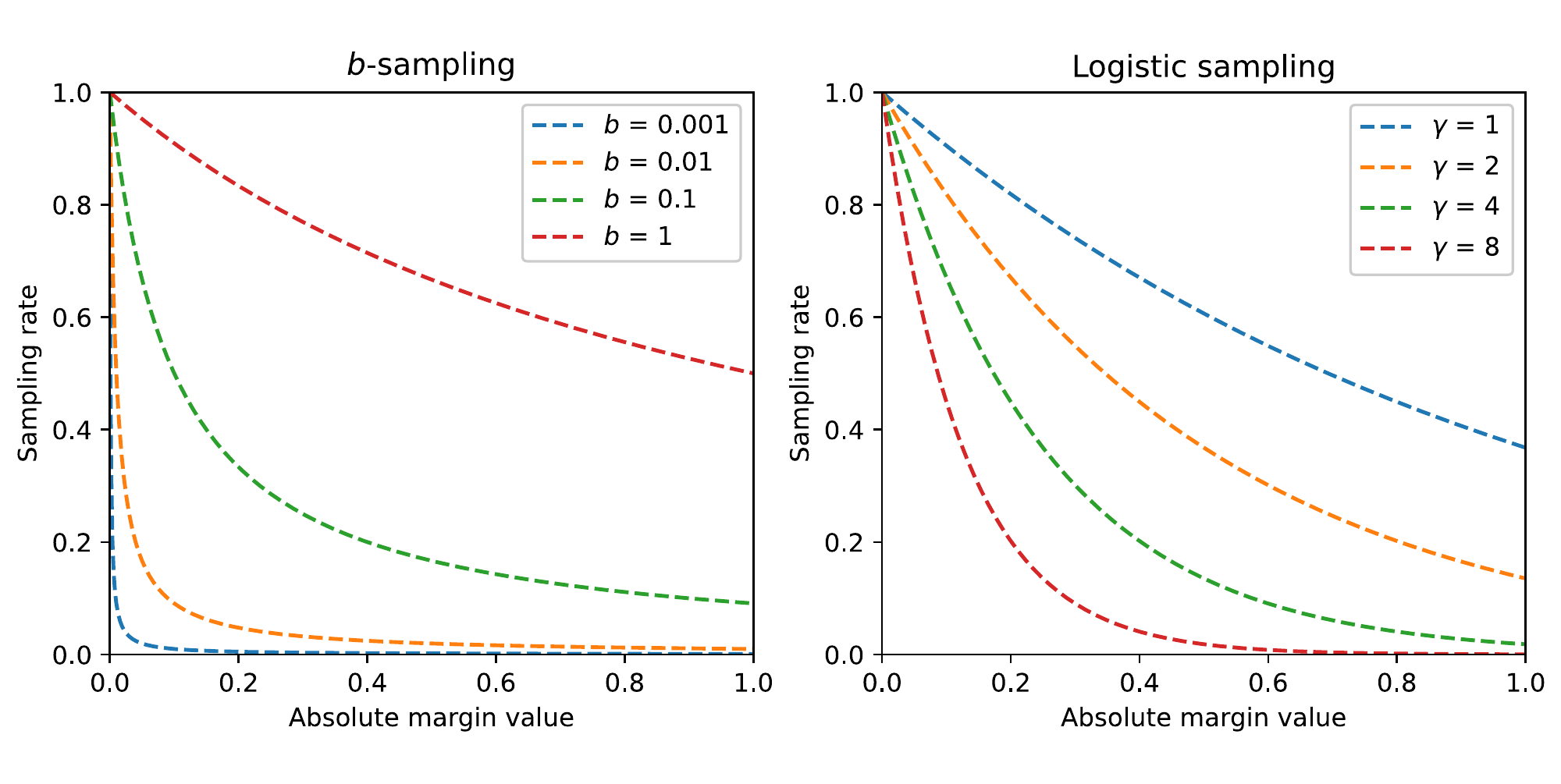}
  \caption{Shape of the sampling distributions for $b$-sampling (a) and logistic sampling (b), for different values of $b$ and $\gamma$.}
  \label{fig:bernoulli_curves}
\end{figure}

\noindent
As in the traditional $b$-sampling approach introduced by \cite{cesabianchi1}, this sampling strategy depends on the uncertainty, meant as the distance from the prediction hyperplane. The main difference between the two strategies is the shape of the resulting sampling distribution, which can be observed in Figure \ref{fig:bernoulli_curves}.

The selective sampling perceptron approach has also been investigated by \cite{Lu2016}, who proposed an online passive-aggressive active learning variant of the algorithm. Similarly to the b-sampling approach, at each time step $t$, a Bernoulli random variable $Z_t \in\{0,1\}$ is drawn to decide whether to query the label of the current data point $\mathbf{x}_t$ or not. In this case, the parameter of $Z_t$ is given by

\begin{equation}
    P_t=\frac{\delta}{\delta+\left|\widehat{p}_t\right|}
\end{equation}

\noindent
where $\delta \geq 1$ is a smoothing parameter. Besides not allowing the smoothing parameter to assume a value lower than $1$, the sampling distribution is the same as the one governed by the parameter in Equation \ref{eq:bernoulli}. The main difference lies in the passive-aggressive approach used for updating the weight vector. Indeed, while the traditional perceptron update, shown in Equation \ref{eq:perceptron}, only uses misclassified examples to update the model, the passive-aggressive approach updates the weight vector $\mathbf{w} \in \mathds{R}^d$ whenever the current loss $\ell_t\left(\mathbf{w}_{t-1} ;\left(\mathbf{x}_t, y_t\right)\right)$ is nonzero \citep{Crammer2006}. The new parameter $\mathbf{w}_t$ is found using 

\begin{equation}
    \mathbf{w}_t=\mathbf{w}_{t-1}+\tau_t y_t \mathbf{x}_t
\end{equation}

\noindent
where $\tau_t$ represents the step size, and can be computed according to three different policies

\begin{equation}
    \tau_t=\left\{\begin{array}{l}
\ell_t\left(\mathbf{w}_{t-1} ;\left(\mathbf{x}_t, y_t\right)\right) /\left\|\mathbf{x}_t\right\|^2 \\
\min \left(\kappa, \ell_t\left(\mathbf{w}_{t-1} ;\left(\boldsymbol{x}_t, y_t\right)\right) /\left\|\boldsymbol{x}_t\right\|^2\right) \\
\ell_t\left(\mathbf{w}_{t-1} ;\left(\mathbf{x}_t, y_t\right)\right) /\left(\left\|\mathbf{x}_t\right\|^2+1 / 2 \kappa\right)
\end{array}\right.
\end{equation}

\noindent
where $\kappa$ is a penalty cost parameter. Passive-aggressive algorithms are known for their aggressive approach in updating the model, which is motivated by the fact that traditional perceptron updates might waste data points that have been correctly classified but with low prediction confidence. 

A related issue to the update of the weight vector $\mathbf{w}_t$ was emphasized by \cite{Bordes2005}, who noted that always picking the most misclassified example is a reasonable sampling strategy only when the training examples are highly confident. When dealing with noisy labels, this strategy could lead to the selection of misclassified examples or examples lying on the wrong side of the optimal decision boundary. To address this, they suggested a more conservative approach that selects examples for updating $\mathbf{w}_t$ based on a minimax gradient strategy. 

In addition to confidence in the labels of the training examples, confidence in the model itself must be considered when the sampling strategy is based solely on model predictions. \cite{Hao2018} pointed out that a margin-based sampling strategy may be suboptimal when the classifier is not precise, especially in the early rounds of active learning when the model performance may be poor due to limited training feedback, leading to misleading sampling decisions. This issue is also referred to as cold-start active learning \citep{Houlsby2014,Yuan2020,Jin2022}. To address this, \cite{Hao2018} propose considering second-order information in addition to margin value when deciding whether or not to query the label of a data point $\mathbf{x}_t$. In general, first-order online active learning strategies only consider the margin value, while second-order methods also take into account the confidence associated with it. To do this, they assume that the weight vector of the classifier $\mathbf{w} \in \mathds{R}^d$ is distributed as

\begin{equation}
    \mathbf{w} \sim \mathcal{N}(\boldsymbol{\mu}, \mathbf{\Sigma})
\label{eq:weights}
\end{equation}

\noindent
where the values $\mu_i$ and $\Sigma_{i, i}$ encode the model knowledge and confidence in the weight vector for the $i$th feature $w_i$. The covariance between the $i$th and $j$th features is captured by the term $\Sigma_{i, j}$. The smaller the variance associated with the coefficient $w_i$, the more confident the learner is about its mean value $\mu_i$. The objective of the proposed method is to take into account the confidence of the model when updating the model and making the sampling decision. With regards to the model update, when the true label $y_t$ of $\mathbf{x}_t$ is queried, the Gaussian distribution in Equation \ref{eq:weights} is updated by minimizing an objective function based on the Kullback-Leibler divergence \citep{Joyce2011} to ensure the updated model is not too different from the previous one. The sampling decision uses an additional parameter to the margin $\widehat{p}_t$, which is defined as

\begin{equation}
    c_t=\frac{-\eta}{2\left(\frac{1}{\nu_t}+\frac{1}{\gamma}\right)}
\end{equation}

\noindent
where $\eta, \gamma>0$ are two fixed hyper-parameters and $\nu_t$ represents the variance of the margin related to the data point $\mathbf{x}_t$. The intuition is that, when the variance $\nu_t$ is high, the model has not been sufficiently trained on instances similar to $\mathbf{x}_t$, and querying its label would lead to a model improvement. Then, a soft margin-based approach is employed by computing

\begin{equation}
    \rho_t=\left|\widehat{p}_t\right|+c_t
\end{equation}

\noindent
If $\rho_t \leq 0$, the label is always queried as the model is extremely uncertain about the margin. Instead, when $\rho_t > 0$, the model is more confident, and the labeling decision is taken by drawing a Bernoulli random variable of parameter

\begin{equation}
    P_t=\frac{\delta}{\delta+\rho_t}
\end{equation}

\noindent
where $\delta > 0 $ is a smoothing parameter. Finally, \cite{Hao2018} also introduced a cost-sensitive variant of the loss function, for dealing with class-imbalanced applications. For a comprehensive discussion on imbalanced data stream analysis, please see \cite{aguiar2023survey}.

The cold-start issue related to the application of active learning to imbalanced datasets has also been highlighted by \cite{Qin2021}, who used extreme learning machines \citep{Huang2006} and extended the active learning framework initially proposed by \cite{Yu2015} to the multiclass classification scenario. They highlighted the challenge of the lack of instances for certain classes in imbalanced datasets, which can seriously impact the predictive ability of the model for those classes. To address this issue, they propose a sampling strategy that considers both diversity and uncertainty. The diversity is calculated by computing pairwise Manhattan distance between the unlabeled observations. The uncertainty of a data point $\mathbf{x}_t$ is computed by taking the difference between the largest two posterior probabilities as in

\begin{equation}
    \operatorname{margin}\left(\mathbf{x}_t\right)=p\left(y=c_b \mid \mathbf{x}_t\right)-p\left(y=c_{s b} \mid \mathbf{x}_t\right)
\label{eq:best-secondbest}
\end{equation}

\noindent
where $c_b$ and $c_{sb}$ are the classes with the highest posterior probabilities. This approach is also referred to as best-versus-second-best margin and, as highlighted by \cite{Joshi2009}, is a good indicator of uncertainty when a large number of classes are present in the data. It should be noted that the sampling strategy introduced by \cite{Qin2021} is not suited for single-pass active learning as it requires computing similarity and uncertainty measures for all the unlabeled observations in the current batch. 

Another approach to deal with class imbalance in active learning was proposed by \cite{Ferdowsi2013}, who used linear SVMs and a sampling strategy that switches between multiple instance selection criteria online. This approach, however, is limited to a pool-based setting and requires predicting an unsupervised evaluation score for all available unlabeled instances. The impact of the last queried observations on the scores associated with the unlabeled data points is evaluated, and a greedy approach is used to decide which instance selection criterion to trust. SVMs have also been used by \cite{Ghassemi2016}, who proposed a differentially private approach to online active learning. The privacy concerns are tackled both during the instance selection and the training phase, by randomizing the strategy introduced by \cite{Tong2002}. The informativeness of a data point $\mathbf{x}_t$ is measured by its closeness to the current hyperplane $\mathbf{w}_t$ as in

\begin{equation}
    c(t)=\exp \left(-d\left(\mathbf{x}_t, \mathbf{w}_t\right)\right) \in[0,1]
\end{equation}

\noindent
where the distance function $d\left(\mathbf{x}_t, \mathbf{w}_t\right)$ is defined as

\begin{equation}
    d\left(\mathbf{x}_t, \mathbf{w}_t\right) \triangleq \frac{\left|\left\langle\mathbf{w}_t, \mathbf{x}_t\right\rangle\right|}{\left\|\mathbf{w}_t\right\|}
\end{equation}

\noindent
In the traditional framework, the label $y_t$ is queried if we have $c(t)>\Gamma$, where $\Gamma$ is a pre-defined threshold. It should be noted that $c(t)>\Gamma$ is equivalent to $d\left(\mathbf{x}_t, \mathbf{w}_t\right) \leq \log 1 / \Gamma$, which means that the observation $\mathbf{x}_t$ is in a sampling region of width $2 \log 1 / \Gamma$ around $\mathbf{w}_t$. However, to avoid a deterministic decision process on the labeling and ensure privacy, some randomness needs to be introduced. This can be done in two ways. First, the labeling decision can be modeled as a Bernoulli random variable of parameter $p$ if $c(t)<\Gamma$ or $(1-p)$ if  $c(t) \geq \Gamma$, where $p<1/2$. Another approach is based on the exponential mechanism introduced by \cite{McSherry2007}. According to this strategy, the algorithm sets a constant probability of labeling data points within a sampling region defined by $\alpha$, and a decaying probability for points outside of it. The selection strategy is represented by a Bernoulli of parameter

\begin{equation}
    q(t)= \begin{cases}e^{-\alpha \epsilon / \Delta} & d\left(\mathbf{x}_t, \mathbf{w}_t\right) \leq \alpha \\ e^{-d\left(\mathbf{x}_t, \mathbf{w}_t\right) \epsilon / \Delta} & d\left(\mathbf{x}_t, \mathbf{w}_t\right)>\alpha\end{cases}
\end{equation}

\noindent
where $\epsilon>0$ and $\Delta=(1-\alpha / M) M$. The authors assumed all data points belonging to the stream to be bounded in norm by $M$, $\left\|\mathbf{x}_t\right\| \leq M \text { for } t=1, \ldots, T$. To tackle the privacy concerns while training, the authors propose two mini-batch strategies, to avoid the problem of slow convergence that may result from introducing noise according to the private stochastic gradient descent scheme \citep{Bassily2014,Song2013,Duchi2013}.

Two different approaches have been proposed by \cite{Ma2016} and \cite{Shah2020}. \cite{Ma2016} proposed a query-while-learning strategy for decision tree classifiers. They used entropy intervals extracted from the evidential likelihood to determine the dominant attributes, which are ordered based on the information gain ratio. When a new data point $\mathbf{x}_t$ is observed, its label is queried only if there does not exist a dominant attribute. This will help to identify one and narrow the entropy interval. However, it should be noted that the authors consider a query while learning framework that only partially relates to to online active learning.  \cite{Shah2020} investigated the online active learning problem for reject option classifiers. Given the high cost that is sometimes associated with a misclassification error, these models are given the option of not predicting anything, for example when dealing with a highly ambiguous instance. A typical application of reject option classifiers is in the medical field, when making a diagnosis with ambiguous symptoms might be particularly difficult. In this case, it could be more beneficial not to provide a prediction but suggest further tests instead. They proposed an approach based on a non-convex double ramp loss function $\ell_{d r}$ \citep{Manwani2013}, where the label of the current example $\mathbf{x}_t$ is queried only if it falls in the linear region of the loss given by $\left|f_t\left(\mathbf{x}_t\right)\right| \in\left[\rho_t-1, \rho_t+1\right]$, which is the region where the parameter would be updated. Here, $\rho$ refers to the bandwidth parameter of the reject option classifier that determines the rejection region.

\cite{fujikashima} investigated the problem of Bayesian online active learning. They provided a general framework based on policy-adaptive submodularity to handle data streams in an online setting. The authors distinguish between the stream setting, where the labeling decision can be made within a given timeframe, and the secretary setting, introduced in Section \ref{sec:preliminaries}, where the labeling decision must be made immediately. The proposed framework can be applied in a variety of active learning scenarios, such as active classification, active clustering, and active feature selection. The framework is based on the concept of adaptive submodular maximization, which extends the idea of submodular maximization. A set function is considered to be submodular if it satisfies the property of diminishing returns, meaning that adding an element to a smaller set has a greater impact on the function value than adding the same element to a larger set. Adaptive submodular maximization allows the model to adapt to the changing distribution of data over time, by adjusting the set function to reflect the current state of knowledge. This leads to more efficient use of available data and improved performance.

So far, we discussed several single model approaches to active learning, which have shown promising results in various applications. However, it is important to note that single models have their limitations and can sometimes struggle to capture complex patterns and diverse representations present in the data. To address these limitations, researchers have proposed the use of ensembles or committees as an alternative \citep{krawczyk2017ensemble}. An ensemble or committee refers to a group of multiple models that collaborate to produce a more robust and accurate prediction by combining their individual predictions. The models in an ensemble or committee can be trained on different subsets of the data or with varying hyperparameters, and the final prediction is typically made through either voting or weighted averaging. Ensembles or committees can also be regarded as a collection of models that work together to make a prediction, either by exchanging information or learning from one another. Among this class of methods, a common sampling strategy is represented by disagreement-based active learning. A framework to perform disagreement-based active learning in online settings was recently introduced by \cite{Huang2022}. They characterized the learner by a hypothesis space $\mathcal{H}$ of Vapnik-Chervonenkis (VC) dimension $d$, which is composed of all the classifiers currently under consideration, and a Tsybakov noise model \citep{Mammen1999,Tsybakov2004}. Each classifier $h \in \mathcal{H}$ is a measurable function mapping the observation $\mathbf{x}_t$ to binary output $y_t=\{0,1\}$. The disagreement among two classifiers is given by $d\left(h_1, h_2\right)=\mathds{P}\left[h_1(\mathbf{x}) \neq h_2(\mathbf{x})\right]$ and the disagreement region is defined as

\begin{equation}
    D\left(h_1, h_2\right)=\left\{\mathbf{x} \in \mathcal{X}: h_1(\mathbf{x}) \neq h_2(\mathbf{x})\right\}
\end{equation}

\noindent
The online active learning strategy is represented by the policy $\pi=\left(\left\{v_t\right\},\left\{\lambda_t\right\}\right)$, where $\left\{v_t\right\}_{t \geq 1}$ is the map of the queried data points, and $\left\{\lambda_t\right\}_{t \geq 1}$ is the sequence of prediction rules. Over the time horizon $T$, the performance of the policy $\pi$ is evaluated using the label complexity and the regret. The label complexity measures the expected number of labels queried, with respect to the stochastic process induced by $\pi$, and it is given by

\begin{equation}
    \mathds{E}[Q(T)]=\mathds{E}\left[\sum_{t=1}^T \mathds{1}\left[v_t=1\right]\right]
\end{equation}

\noindent
The regret, on the other hand, represents the expected number of excess classification errors with respect to $h^*$, and it is obtained as

\begin{equation}
    \mathds{E}[R(T)]=\mathds{E}\left[\sum_{t \leq T: v_t=0} \mathds{1}\left[\lambda_t \neq y_t\right]-\mathds{1}\left[h^*\left(\mathbf{x}_t\right) \neq y_t\right]\right]
\end{equation}

\noindent
The objective of the algorithm is to minimize the label complexity with a constraint on the regret. At the first round, the initial version space is the entire hypothesis space $\mathcal{H}$, while the initial region of disagreement is the whole input space $\mathcal{X}$. Then, at time step $t$, the learner updates the version space $\mathcal{H}_t$ using the $M$ collected labels, and computes a new region of disagreement as

\begin{equation}
    \mathcal{D}\left(\mathcal{H}_t\right)=\left\{\mathbf{x} \in \mathcal{X}: \exists h_1, h_2 \in \mathcal{H}_t, h_1(\mathbf{x}) \neq h_2(\mathbf{x})\right\}
\end{equation}

\noindent
If $\mathbf{x}_t \in \mathcal{D}\left(\mathcal{H}_t\right)$, then the label of the current data point is queried, otherwise a prediction is produced using an arbitrary classifier in $\mathcal{H}_t$. At the end of the iteration $t$, the set $\mathcal{Z}_t$ of $M$ collected labeled examples is used to estimate the empirical error $\epsilon_{\mathcal{Z}_t}(h)$ of the classifiers in $\mathcal{H}$ and identify the best currently available classifier. Then, the version space is updated by removing all the suboptimal hypotheses whose empirical error exceeds the one obtained with $h_t^*$ by a threshold $\Delta_{\mathcal{Z}_t}\left(h, h_t^*\right)$.  The threshold regulates the trade-off between reducing label complexity by narrowing the region of disagreement and increasing the regret by eliminating good classifiers. 

The disagreement concept was also used by \cite{Desalvo2021}, while proposing an approach to online active learning for binary classification tasks based on surrogate losses. The overall framework is similar to the disagreement-based one used by \cite{Huang2022}, with the main difference being the use of weak-labels to optimize the sampling strategy. At each time step $t$, the learner observes the unlabeled data point $\mathbf{x}_t$ and either decides to request its label or assigns a pseudo-label $\widehat{y}_t$. Then, the pseudo labels $\widehat{y}_t$ and the true labels $y_t$ processed so far are used together to obtain an estimate of the empirical risk $\epsilon_{\mathcal{S}_t}(h)$, where $\mathcal{S}_t$ is obtained by combining the collected labeled examples $\mathcal{Z}_t$ with the pseudo-labeled ones $\widehat{\mathcal{Z}}_t$. This represents an example of combining active learning and semi-supervised learning, as highlighted in Section \ref{subsec:ssl}.

\cite{Loy2012} presented a Bayesian framework that leverages the principle of committee consensus to balance exploration and exploitation in online active learning. The aim of exploration is to discover new, previously unknown classes, while exploitation focuses on refining the decision boundary for known classes. To address the issue of unknown classes, the framework uses a Pitman-Yor Processes (PYP) prior model \citep{Pitman1997} with a Dirichlet process mixture model (DPMM). A DPMM is a non-parametric clustering and classification model that models the data generating process using a mixture of probability distributions. Each data point is assigned to a cluster, which is associated with a probability distribution over the classes. The number of clusters is modeled using a Dirichlet process, which is a distribution over distributions that allows for an infinite number of clusters but ensures that the number of actual clusters is always finite. At each time step $t$, the learner samples two random hypotheses $h_1$ and $h_2$ from the model. Then, it computes the posterior probability of the current class $c$ corresponding to $k$, $p\left(c=k \mid \mathbf{x}_t\right)$, for each of the two hypotheses. Finally, $h_i\left(\mathbf{x}_t\right)=\arg \max p\left(c \mid \mathbf{x}_t\right)$  is calculated for $i=1,2$. The label of the current data point is queried in two cases: first, if $h_1\left(\mathbf{x}_t\right) \neq h_2\left(\mathbf{x}_t\right)$, meaning the two hypotheses disagree, and second, if $h_i\left(\mathbf{x}_t\right)=K+1 \forall i$, where $K$ is the number of currently known classes, meaning the current data point belongs to a new class. 

The DPMM has also been used by \cite{Mohamad2020}, who proposed a semi-supervised strategy for performing active learning in online human activity recognition with sensory data. To account for the possibility of dealing with different sensor network layouts, the authors proposed pre-training a conditional restricted Boltzmann machine \citep{Taylor2009,Taylor2006} and used it to extract generic features from the sensory input. The instance selection strategy follows a Bayesian approach, in trying to minimize the uncertainty about the model parameters. To assess the usefulness of labeling the data point $\mathbf{x}_t$, they measure the discrepancy between the model uncertainty computed from the data observed until the time step $t$ and the expected risk associated with $y_t$. This gives a hint of how the current label would impact the current model uncertainty. A dynamically adaptive threshold $\Gamma$ is finally used to the determine whether the current expected risk is greater than the current risk.

A different kind of committee has been considered by \cite{Hao2018Expert}. They proposed a framework for minimizing the number of queries made by an online learner that is trying to make the best possible forecast, given the advice received from a pool of experts. To do so, they adapted the exponentially weighted average forecaster (EWAF) and the greedy forecaster (GF) to the online active learning scenario. A comprehensive analysis of forecasters to perform prediction with expert advice can be found in the book by \cite{plg}. In general, at each time step $t$, the learner or forecaster has access to the predictions for the data point $\mathbf{x}_t$ made by the $N$ experts, $f_{i, t}\left(\mathbf{x}_t\right): \mathds{R}^d \rightarrow[0,1] \text { with } i=1, \ldots, N$. Based on these predictions, it outputs its own prediction $p_t$ for the outcome $y_t$. Then, if the label is revealed, the predictions made by the forecaster and the experts are scored using a nonnegative loss function $\ell$. The objective of the learner is to minimize the cumulative regret over the time horizon $T$, which can be seen as the difference between its loss and the one obtained with each expert $i$ as in

\begin{equation}
    R_{i, T}=\sum_{t=1}^T\left(\ell\left(p_t, y_t\right)-\ell\left(f_{i, t}\left(\mathbf{x}_t\right), y_t\right)\right)=\widehat{L}_T-L_{i, T}
\end{equation}

\noindent
The most simple approach to obtain a prediction $p_t$ from the learner is to compute a weighted average of the experts predictions as in

\begin{equation}
    p_t=\frac{\sum_{i=1}^N \omega_{i, t} f_{i, t}\left(\mathbf{x}_t\right)}{\sum_{i=1}^N \omega_{i, t}}
\end{equation}

\noindent
where $\omega_{i, t}\geq 0$ is the weight assigned at time $t$ to the $i$th expert. 
With the EWAF, the weight for the $i$th expert are obtained using

\begin{equation}
    \omega_{i, t}=\frac{e^{\eta R_{i, t-1}}}{\sum_{i=1}^N e^{\eta R_{i, t-1}}}
\end{equation}

\noindent
where $\eta$ is a positive decay factor and $R_{i, t-1}$ is the cumulative loss of expert $i$ observed until step $t$. The exponential decay factor $\eta$ determines the weight given to the past losses, with more recent losses having a higher weight and older losses having a lower weight. Instead, the GF works by minimizing, at each time step, the largest possible increase of the potential function for all the possible outcomes of $y_t$. The potential function is the function that assigns a potential value to each expert, which captures the quality of an expert advice based on its past performance. 
\cite{Hao2018Expert} extended the EWAF and GF by proposing the active EWAF (AEWAF) and active GF (AGF). The key idea is that, while the standard EWAF and GF assume the availability of the true label $y_t$ after each prediction, in the online active learning framework the loss $\ell$ can only be measured a limited number of times. To factor this in, a binary variable $Z_t \in\{0,1\}$ is introduced to decide whether or not at round $t$ the label is requested. Consequently, the cumulative loss suffered by the $i$th expert on the instances queried by the active forecaster is given by

\begin{equation}
    \widehat{L}_{i, T}=\sum_{t=1}^T \ell\left(f_{i, t}\left(\mathbf{x}_t\right), y_t\right) \cdot Z_t
\end{equation}

\noindent
The sampling strategy is based on the determination of a confidence condition on the difference between the prediction $p_t$ of the fully supervised forecaster and the prediction $\widehat{p}_t$ made by the active forecaster. For the active forecaster we have that $\widehat{p}_t=\pi_{[0,1]}\left(\overline{p}_t\right)$, where $\overline{p}_t$ depends on the chosen model. The AEWAF is based upon the observation that if we have

\begin{equation}
    \max _{1 \leq i, j \leq N}\left|f_{i, t}\left(\mathbf{x}_t\right)-f_{j, t}\left(\mathbf{x}_t\right)\right| \leq \delta
\label{eq:cond1}
\end{equation}

\noindent
then $\left|p_t-\widehat{p}_t\right| \leq \delta$, where $\delta$ is a tolerance threshold. This means that the prediction of the forecaster is close to the one obtained in the fully supervised setting if the maximum difference of advice between any two experts is not too large. This assumption might not hold in the presence of noisy or bad experts and, to tackle this problem, the authors proposed a robust variant of the AEWAF. The AGF uses instead a confidence condition based on the fact that if

\begin{equation}
    \max _{1 \leq i, j \leq N}\left|f_{i, t}\left(\mathbf{x}_t\right)-\overline{p}_t\right| \leq \delta
\label{eq:cond2}
\end{equation}

\noindent
then $\left|p_t-\widehat{p}_t\right| \leq \delta$. The general scheme for performing online active learning with expert advice is reported in Algorithm \ref{alg:2}.

\begin{algorithm}
\caption{Online active learning with expert advice}\label{alg:2}
\begin{algorithmic}
\Require a data stream $\mathbf{S}$, a loss function $\ell$, a time horizon $T$, a set of $N$ experts, a tolerance threshold $\delta$, a sampling budget $B$.
\State $t \gets 1$ \Comment{Timestamp}
\State $c \gets 0$ \Comment{Labeling cost}
\While{$c \leq B$, $t \leq T$}
    \State Observe an incoming data point $\mathbf{x}_t \in \mathcal{S}$
    \State Receive advide by experts $\left\{f_{i, t}\left(\mathbf{x}_t\right): i=1, \ldots, N\right\}$
    \State Generate prediction $\overline{p}_t$ for the label $y_t$ and set $\widehat{p}_t=\pi_{[0,1]}\left(\overline{p}_t\right)$
    \State Draw a Bernoulli random variable $Z_t$ of parameter $P_t=b /\left(b+\left|\widehat{p}_t\right|\right)$
    \If {Equation \ref{eq:cond1} or \ref{eq:cond2} is satisfied } \Comment{Sampling decision}
        \State Discard $\mathbf{x}_t$
    \Else
        \State Ask for the true label $y_t$
        \State $c \gets c + 1$ \Comment{Pay for the label}
    \EndIf
    \State $t \gets t + 1$
\EndWhile
\end{algorithmic}
\end{algorithm}

A similar framework, in conjunction with multiple kernel learning (MKL), has been investigated by \cite{Chae2021}. They propose an active MKL (AMKL) algorithm based on random feature approximation. In general, online MKL based on random feature approximation is a method for online learning and prediction that combines multiple kernel functions to improve the performance of a learning algorithm \citep{Jin2010,Hoi2013}. In MKL, multiple kernel functions are used to capture different aspects of the data, and the optimal combination of kernels is learned from the data. The online version of MKL based on random feature approximation is designed to handle data that arrives sequentially, and the learning algorithm is updated after each new data point. In kernel-based learning, the target function $f(\mathbf{x})$ is assumed to belong to a reproducing Hilbert kernel space (RKHS). In the proposed AMKL the learner uses an ensemble of $N$ kernel functions. At each time step $t$, two main steps are implemented. First, each kernel function $\hat{f}_{i, t}\left(\mathbf{x}_t\right), \text { with } i=1, \ldots, n$, is optimized independently of the other kernel functions. This is referred to as local step. Then, in the global step, the learner seeks the best function approximation $\widehat{f}_t\left(\mathbf{x}_t\right)$ by combining the $N$ kernel functions as in

\begin{equation}
    \widehat{f}_t\left(\mathbf{x}_t\right)=\sum^N \widehat{v}_{i, t} \hat{f}_{i, t}\left(\mathbf{x}_t\right)
\end{equation}

\noindent
where $\widehat{v}_{i, t}$ refers to the weight for the $i$th kernel function at round $t$. Similarly to the case with expert advice, the weights are determined by minimizing the regret over the time horizon $T$, which is defined as the difference between the loss of the learner and the one obtained with the best kernel function $f_{i, t}^*$. To do so, the weights are computed based on the past losses $\ell$ as

\begin{equation}
    \widehat{\omega}_{i, t}=\exp \left(-\eta_g \sum_{\tau \in \mathcal{A}_{t-1}} \ell\left(\hat{f}_{i, \tau}\left(\mathbf{x}_\tau\right), y_\tau\right)\right)
\end{equation}

\noindent
where $\eta_g>0$ is a tunable parameter and $\mathcal{A}_{t-1}$ is an index of time stamps $t$ indicating the instances for which has label has been requested, thus permitting to measure the loss. Then, the weights $\widehat{v}_{i, t}$ are obtained from $\widehat{\omega}_{i, t}$ as follows

\begin{equation}
    \widehat{v}_{i, t}=\frac{\widehat{\omega}_{i, t}}{\sum_{i=1}^N \widehat{\omega}_{i, t}}
\end{equation}

\noindent
Finally, the instance selection criterion is based on a confidence condition, denoted by with $\delta > 0$, on the similarity of the learned kernel function, which is a similar to the condition used by \cite{Hao2018Expert} in the formulation of the AEWAF

\begin{equation}
    \max _{1 \leq j \leq N} \sum_{i=1}^N \widehat{v}_{i, t} \ell\left(\widehat{f}_{i, t}\left(\mathbf{x}_t\right), \widehat{f}_{j, t}\left(\mathbf{x}_t\right)\right) \leq \delta
\end{equation}

\subsection{Drifting data stream classification approaches} \label{subsec:drifting}
Active learning strategies belonging to this category aim to tackle online classification tasks in time-varying data streams affected by distribution shifts. We can classify distribution shifts into three main categories, depending on whether they concern the feature space $\mathbf{x}$ or the output dimension $y$. A shift that only affects the input distribution $p(\mathbf{x})$, and not the conditional distribution $p(y \mid \mathbf{x})$, is referred to as covariate shift \citep{Zhou2021,Wu2021,Li2021BayesianOnlineLearning} or virtual drift \citep{Baier2021}. In these circumstances, for two different time steps, $t_i$ and $t_{i+\Delta}$, we have that $p_{t_i}(\mathbf{x}) \neq p_{t_{i+\Delta}}(\mathbf{x})$ and $p_{t_i}(y \mid \mathbf{x}) = p_{t_{i+\Delta}}(y \mid \mathbf{x})$, meaning that the underlying model is not being altered by phenomena like class swaps or coefficient changes. Conversely, in the presence of a real concept drift \citep{Baier2021,quintana}, the conditional distribution changes, and we have $p_{t_i}(y \mid \mathbf{x}) \neq p_{t_{i+\Delta}}(y \mid \mathbf{x})$. In this scenario, the predictive performance of the fitted model dramatically deteriorates, and a model update or replacement becomes necessary. An example of this kind of distribution shift can be identified in the changes of the consumer behaviors over time, or following a major event as the COVID-19 pandemic \citep{Zwanka2021}. However, it should be noted that virtual drifts and real concept drifts often occur together \citep{Tsymbal2008}, leading to a situation where we have both $p_{t_i}(\mathbf{x}) \neq p_{t_{i+\Delta}}(\mathbf{x})$ and $p_{t_i}(y \mid \mathbf{x}) \neq p_{t_{i+\Delta}}(y \mid \mathbf{x})$ \citep{Lu2018}. Lastly, we can incur in a label distribution shift \citep{Wu2021} when the shift only affects $p(y)$, leading to $p_{t_i}(y) \neq p_{t_{i+\Delta}}(y)$. This situation can be observed in many real-world scenarios where the target distribution changes over time. A typical example is the prediction of diseases like influenza, whose distribution can dramatically change depending on the season, or in the presence of sudden outbreaks. 

\begin{figure}[h]
  \centering
  \includegraphics[width=\linewidth]{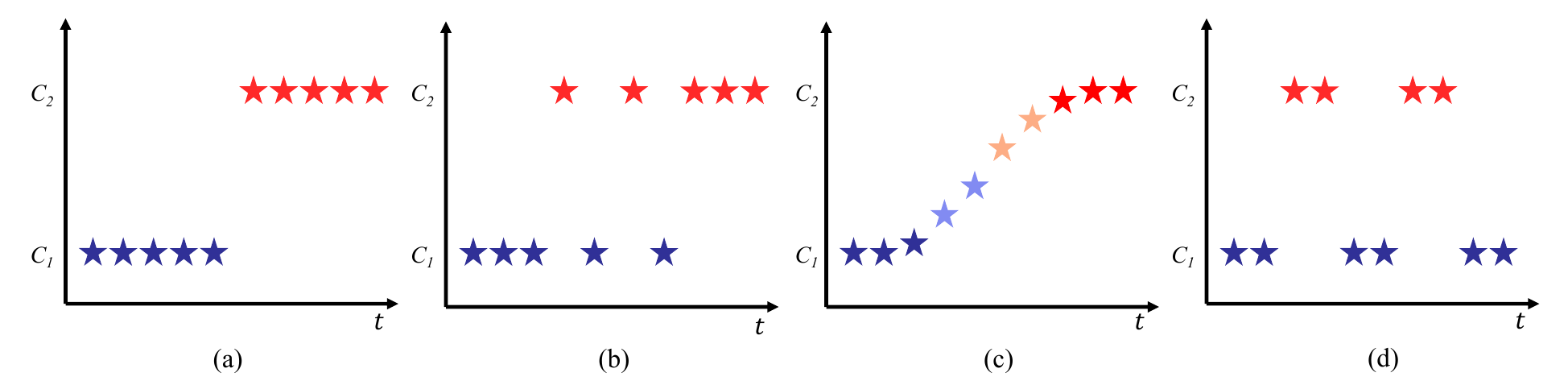}
  \caption{Different types of drifts that can affect the data stream: abrupt drift (a), gradual drift (b), incremental drift (c), recurring concepts (d). $C_1$ and $C_2$ indicate the two concepts that might characterize the data distribution.}
  \label{fig:drifts}
\end{figure}

Another key characteristic of distribution shifts is represented by the change rate, namely how fast the new concept or distribution is introduced into the data stream. To this extent, we can identify four kinds of drifts \citep{Lu2018,Lima2022}, which are illustrated in Figure \ref{fig:drifts}. A sudden or abrupt drift is a drift that can be immediately detected from two consecutive time steps, $t_i$ and $t_{i+1}$. It refers to a sudden and clearly identifiable change in the data distribution. An example of this would be a sudden change in the weather, which would affect the behavior of customers at a retail store. The change is noticeable, and the model needs to be updated immediately. A gradual drift exhibits a transition phase, where a mixture or overlap between the two distributions $p_{t_i}$ and $p_{t_{i+\Delta}}$ exists. In this case, the change is slower and more difficult to detect, making it challenging to update the model. An example would be a change in consumer behavior over time, which is hard to detect but can have a significant impact on a business. Another type of drift is the incremental drift, which has an extremely low transition rate, which makes it very difficult to detect changes between the data points observed in the transition period. This type of drift is often caused by changes in the data generating process that happen gradually over time, in small steps rather than all at once. An example would be changes in the types of products that are popular among customers, which happen gradually and are hard to detect. Finally, a data stream can also be affected by recurring concepts, which sequentially alternate over time. An example would be a retail store where the same types of products are popular at different times of the year, such as winter coats and summer dresses. The model needs to be able to detect and adapt to these recurring concepts in order to maintain good performance. 

In online active learning for drifting data streams, some approaches address the presence of concept drifts by combining active learning strategies with drift detectors \citep{Zhang2020,Krawczyk2018}. Drift detectors are algorithms that try to detect distribution shifts and identify when the context is changing. They can be divided into three macro-categories \citep{Lu2018}. The first group of methods is represented by the error-based drift detectors, which try to detect online changes in the error rate of a base classifier. Among these, one of the most commonly employed strategies is the drift detection method (DDM) proposed by \cite{Gama2004}. Another popular approach is the adaptive window (ADWIN) strategy proposed by \cite{Bifet2007}. The second class of drift detectors is called data distribution-based drift detection, and the third class is represented by multiple hypothesis testing strategies. While the first class contains the majority of the proposed approaches, it assumes that we are able to observe the labels of all the incoming data points to assess the error rate. Instead, the last two classes could be implemented even in an unsupervised manner. An exhaustive overview on unsupervised drift detection methods has been proposed by \cite{Gemaque2020}. While the unsupervised nature of the data distribution-based and multiple hypothesis testing strategies make them ideal for the active learning scenario, it should be noted that real concept drifts can hardly be detected in a completely unsupervised fashion. Indeed, in a circumstance when the input distribution $p(\mathbf{x})$ remains unaltered while the underlying model relating the input variables $\mathbf{x}$ to the label $y$ changes, it would not be possible to detect the change of concept without collecting labels. This is why \cite{Krawczyk2018} propose to apply an error-based drift detector to the few labels collected during the online active learning routine. To this extent, they use the ADWIN \citep{Bifet2007} method to detect drifts and decide when the current model needs to be updated or replaced. The proposed general framework for dealing with online active learning with drifting data streams is reported in Algorithm \ref{alg:3}.

\begin{algorithm}[h]
\caption{Online active learning with drifting data streams}\label{alg:3}
\begin{algorithmic}[h]
\Require a data stream $\mathbf{S}$, a classifier $\Theta$, a drift detector $\Theta$, a sampling strategy $\Upsilon$, a labeling rate $\alpha$, a sampling budget $B$.
\State $t \gets 1$ \Comment{Timestamp}
\State $c \gets 0$ \Comment{Labeling cost}
\While{$c \leq B$ and $t \leq |S|$}
    \State Observe incoming data point $x_t \in \mathbf{S}$
    \If{$\Upsilon(x_t) = \texttt{True}$} \Comment{Sampling decision}
        \State Ask for the true label $y_t$
        \State $c \gets c+1$ \Comment{Pay for the label}
        \State Update classifier $\Psi$ with the labeled example $\left(\mathbf{x}_t, y_t\right)$
        \State Update drift detector $\Theta$ with the labeled example $\left(\mathbf{x}_t, y_t\right)$
        \If{\texttt{drift warning} $=\texttt{True}$}
            \State Start to train a new classifier $\Psi_\text{new}$
            \State Increase labeling rate $\alpha$
        \Else
            \If{\texttt{drift detected} $=\texttt{True}$} \Comment{A detection is always preceded by a warning}
                \State Replace $C$ with $C_\text{new}$
                \State Further increase $\alpha$
            \Else
                \State Return to initial labeling rate $\alpha$
            \EndIf
        \EndIf
    \If{$C_\text{new}$ exists} \Comment{Keeps being updated in the background until replacement}
        \State Update classifier $C_\text{new}$ with the labeled example $\left(\mathbf{x}_t, y_t\right)$
    \EndIf
\EndIf
\State $t \gets t+1$
\EndWhile
\end{algorithmic}
\end{algorithm}

\begin{figure}[h]
  \centering
  \includegraphics[width=0.5\linewidth]{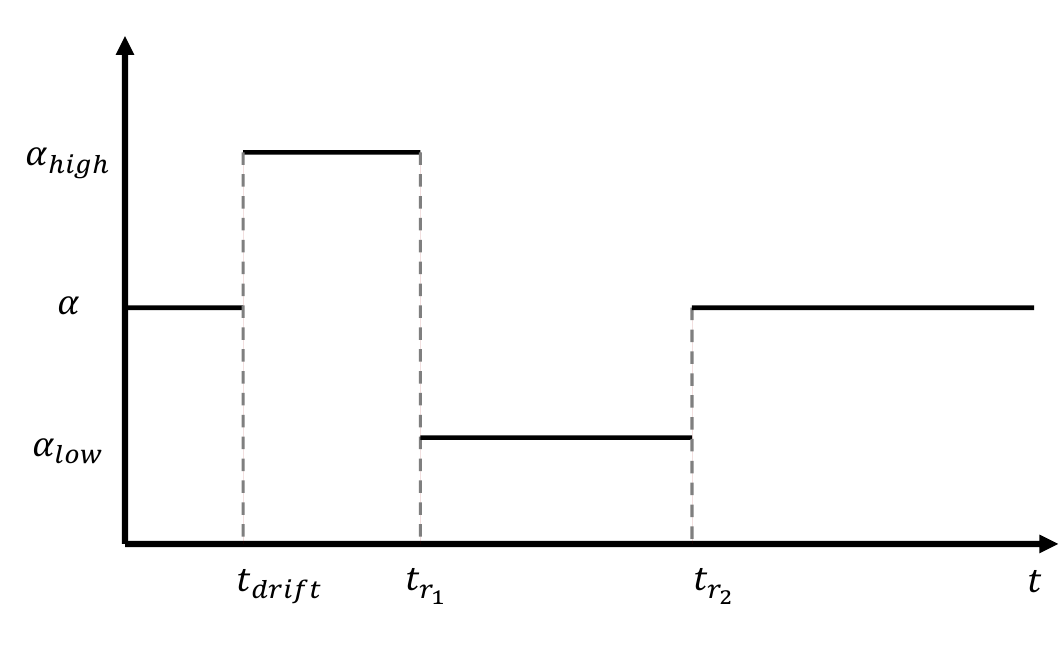}
  \caption{Piece-wise constant budget function introduced by \cite{Castellani2022}. The sampling rate $\alpha$ is increased to $\alpha_{high}$ when a drift is detected ($t_{drift}$), then reduced to $\alpha_{low}$ between $t_{r_1}$ and $t_{r_2}$, before being restored to its nominal value.}
  \label{fig:budget}
\end{figure}

Moreover, the authors proposed the use of a time-variable threshold to balance the budget use over time. Their approach is based on the intuition that, when a new concept is introduced, more labeling effort will be required to quickly collect representative observations belonging to the new concept and replace the outdated model. This is obtained by adjusting a time-variable threshold to balance the budget use over time. Given a threshold $\Gamma$ on the uncertainty of the classifier and a labeling rate adjustment $r \in[0,1]$, the threshold is reduced to $\Gamma - r$ when ADWIN raises a warning and to $\Gamma - 2r$ when a real drift is detected. Thus, when allocating the labeling budget, the key requirement is that the labeling rate employed when a drift is detected should be strictly larger than the one used in static conditions. A similar thresholding idea has also been used by \cite{Castellani2022}, who proposed an active learning strategy for non-stationary data streams in the presence of verification latency. They used a piece-wise constant budget function, where the labeling rate $\alpha$ is increased to $\alpha_{high}$ when a drift is detected and, after a while, reduced to $\alpha_{low}$. Finally, the labeling rate is restored to its nominal value $\alpha$. A visual representation of the labeling approach is shown in Figure \ref{fig:budget}. The length of the time segments where the labeling rate is altered depends on the desired values for $\alpha_{high}$ and $\alpha_{low}$, constraining the overall labeling rate to be equal to $\alpha$.

The authors also tackled the verification latency issue by considering the spatial information of a queried point for which the label has not been made available yet by the oracle. In this way, it is possible to avoid oversampling from regions where many close points have a high utility, namely a low classification confidence. While assessing the utility of the incoming data points the authors use real and pseudo-labels by propagating the information contained in the already labeled observations, as suggested by \cite{Pham2022}. The idea is to use the spatial information of the queried labels by estimating the still missing labels with a weighted majority vote of the label of its k-nearest neighbors labels, where the weight for each nearest neighbor depends on the arrival time of the labels. The verification latency issue in online active learning with drifting data streams was also extensively analyzed by \cite{Pham2022}. Consider the general case where at time $t_i^x$ we draw an instance $\mathbf{x}_i$, and find it interesting enough to send it to the oracle, which will send back the label $y_i$ only at time $t_i^y$, where $t_i^y>t_i^x$. Before the requested label arrives, we might encounter another instance similar to $\mathbf{x}_i$ and ask again for its label, since the learner could not update its utility function or threshold. Similarly, we might use outdated information when updating the policy in a future window. To tackle these issues, the authors propose a forgetting and simulating strategy to avoid using soon-to-be outdated observations and prevent redundant labeling. The instance selection is based upon the variable uncertainty strategy proposed by \cite{Zliobaite2014} and the balanced incremental quantile filter by \cite{Kottke2015}. If we denote the current sliding window at time $t_n^x \text { as } \mathcal{W}_n=\left[t_n^x-\Delta, t_n^x\right)$ and use windows of fixed size $\Delta$, we know that the sliding window that would be used for training when the label $y_n$ related to $\mathbf{x}_n$ arrives will be given by $\mathcal{D}_n=\left[t_n^y-\Delta, t_n^y\right)$. The forgetting step is then implemented by discarding outdated labeled examples that are included in $\mathcal{W}_n$ but will not be included in $\mathcal{D}_n$. If $a_i$ is a Boolean variable indicating whether the $i$th observation has been labeled, the set of instances selected to be forgotten is given by

\begin{equation}
    O_n=\left[\left(\mathbf{x}_i, y_i\right) \forall i<n: a_i=1 \wedge t_i^x, t_i^y \in \mathcal{W}_n \backslash \mathcal{D}_n\right].
\end{equation}

\noindent
Similarly, there is a second set of observations, with time stamps $\mathcal{D}_n^{+}=\mathcal{D}_n \backslash \mathcal{W}_n=\left[t_n^x, t_n^y\right)$, where there might be instances that have been queried but whose label is not currently available. To avoid losing such information and redundantly asking for the label of similar instances, the algorithm simulates incoming labels with a bagging approach by averaging across multiple utility estimations. They also consider an alternative simulation approach based on fuzzy labeling. 

Similarly to \cite{Krawczyk2018}, the ADWIN drift detector has also been used by \cite{Zhang2020} while proposing a method for dealing with online active learning in environments characterized by concept drifts and class imbalance. The instance selection criterion is based on the predictive uncertainty, which they estimate using the best-versus-second-best margin value (Equation \ref{eq:best-secondbest}), as they tackle a multi-class classification problem. An initial pool of $n$ observations is passively collected from the stream to initialize the active learning strategy. Then, a threshold $\Gamma_i$ is estimated for each class as in 

\begin{equation}
    \Gamma_i= \begin{cases}\frac{n m}{n_i L} & \frac{n}{n_i L} \geq 1 \\ m & \frac{n}{n_i L}<1\end{cases}
\end{equation}

\noindent
where $i=1, \ldots, L$ is the number of classes and $m$ is a pre-defined constant used to control the size of the threshold. The model is represented by an ensemble of N classifiers and, when ADWIN detects a concept drift, the classifier with the higher error is replaced with a newly trained one. Finally, the class imbalance issue is also taken into account in two ways, during the training of the ensemble with the use of class-specific weights, and during the active learning routine, by dynamically adjusting the threshold to select more observations belonging to the minority class. 

Recently, \cite{cheng2023active} presented another approach to combine online active learning with drift detection. Their method involves segmenting the data stream $\mathbf{S}$ into fixed-length chunks and then detecting drifts by comparing the distributions of adjacent chunks. After a drift is detected, a multi-objective optimization problem is formulated in order to identify the most relevant and diverse data points within the current batch. For a data point $\mathbf{x}_t$, relevance is defined as its contribution to the new concept, and diversity as the Pearson correlation coefficient with other instances in the same region. Instead, \cite{martins2023meta} proposed to sample the most uncertain data points from each chunk, using a meta-learning framework to fine-tune the threshold used for each window. This allows to reduce the need for labels while maintaining a steady adaptation to the new concepts.

\begin{figure}[h]
  \centering
  \includegraphics[width=0.7\linewidth]{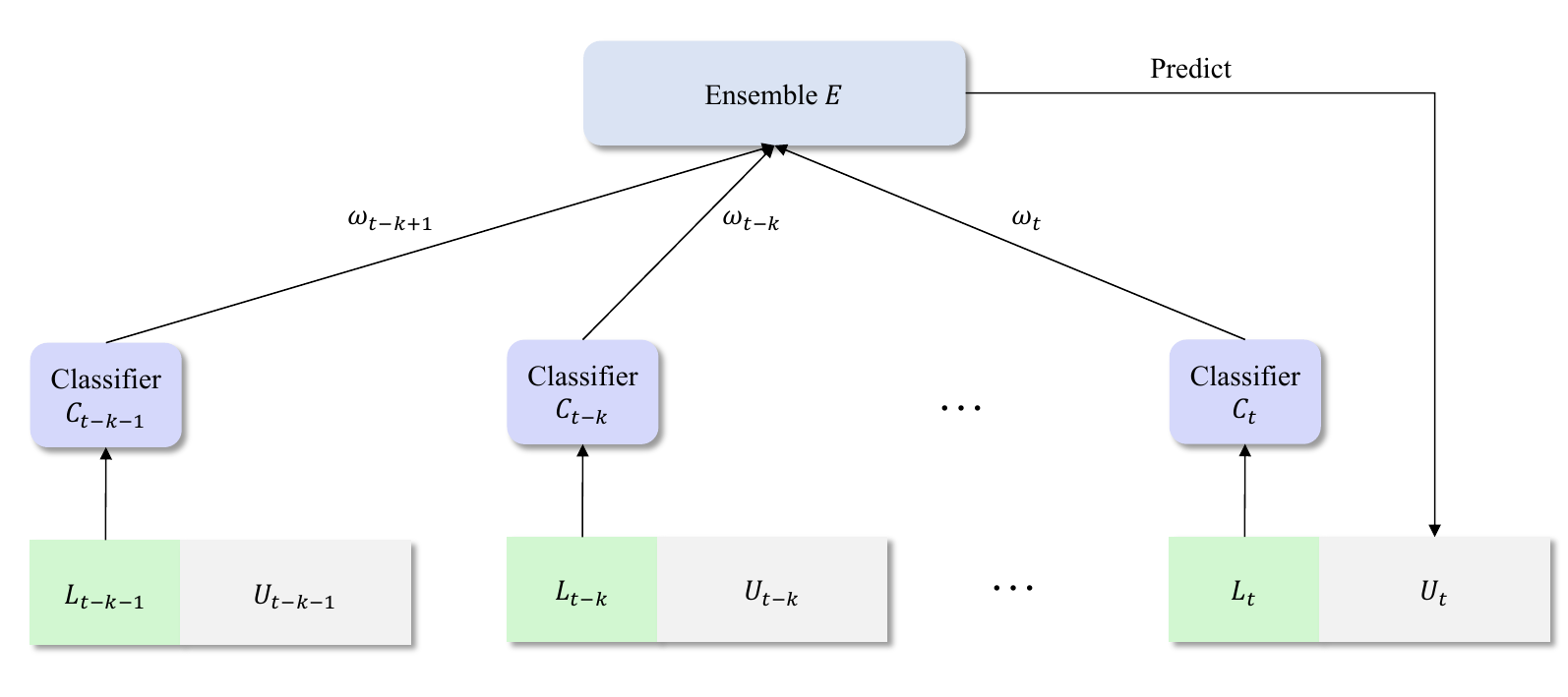}
  \caption{Ensemble-based active learning framework for data streams proposed by \cite{Zhu2007}.}
  \label{fig:ensemble}
\end{figure}

Another window-based approach to perform active learning from data streams has been proposed by \cite{Zhu2007}. The authors developed an ensemble $E$ by partitioning the data stream $\mathbf{S}$ into chunks and then training each of the $k$ models composing the ensemble $E$ on a different chunk of data. In this way, even if the previous observations become unavailable, the models can be used when taking the sampling decision in order to take into account a global uncertainty measure, which is a more robust approach than treating each chunk as a static dataset. At time step $t$, the learner receives a data chunk $\mathbf{S}_t$, which is used to build the current classifier $C_t$. At this point, the ensemble is composed by $C_t$, together with the most recent $k-1$ classifiers, $C_{t-k+1}, \ldots, C_{t-1}$, trained on the labeled examples sampled from the previously observed data chunks, $L_{t-k+1}, \ldots, L_{t-1}$. At each iteration, the objective is to predict the remaining unlabeled data points from the current chunk, $U_t$. The ensemble-based active learning framework is depicted in Figure \ref{fig:ensemble}. The instances selected to be queried are the ones with the largest ensemble variance, and the predictions are obtained by combining the predictions of the single classifiers using the weights $\omega_{t-k+1}, \ldots, \omega_{t-1}$. Finally, a weight updating rule is used to adapt to dynamic data streams.

\cite{Shan2019} and \cite{Zhang2018} developed online active learning strategies by building upon the pairwise classifiers strategy introduced by \cite{Xu2016}. The pairwise strategy makes use of two models, a stable classifier $C_s$ and a dynamic classifier $C_d$, and divides the data stream into batches as in \citep{Zhu2007}. The prediction for an incoming data point $\mathbf{x}_t$ is obtained with a weighted average of the predictions obtained from the two classifiers as in

\begin{equation}
    f_E\left(\mathbf{x}_t\right)=\omega_s f_{C_s}\left(\mathbf{x}_t\right)+\omega_d f_{C_d}\left(\mathbf{x}_t\right)
\end{equation}

\noindent
where $\omega_s$ and $\omega_d$ are the weights associated with the stable and the dynamic classifier, respectively. At time t, the stable classifier $C_s$ is trained on the labeled portions of all the batches processed so far, $L_1, \ldots, L_{t-1}$. Conversely, $C_d$ is trained exclusively on $L_{t-1}$. The key idea is that whenever the reactive classifier starts to outperform the stable classifier, the stable classifier is replaced by the reactive one, which is eventually reset. This replacement allows the learner to adapt to the drift and focus on the most recent instances, forgetting the seemingly obsolete data points. The main drawback of this approach is that it cannot effectively address gradual drifts as the replacement with the classifier trained on the most recent observations makes the learner forget about observations away from the current window. Hence, similarly to the approach of \cite{Zhu2007}, \cite{Shan2019} proposed an extension of this approach, based on an ensemble of classifiers in trying to contemporarily address gradual drifts and abrupt drifts. In their strategy, the stable classifier learns from all the labeled instances and the reactive classifier is replaced by an ensemble of dynamic classifiers, trained on multilevel sliding windows to capture changes in the data stream at different time intervals. The instance selection approach combines random sampling and uncertainty sampling, where the latter is based on the margin value of the predictions obtained by the ensemble. It should be noted that the prediction $f_E$ for the data point $\mathbf{x}_t$ is obtained as a weighted combination of the predictions obtained with the stable and dynamic classifiers as in

\begin{equation}
    f_E\left(\mathbf{x}_t\right)=\omega_s f_{C_s}\left(\mathbf{x}_t\right)+\sum_{d=1}^D \omega_d f_{C_d}\left(\mathbf{x}_t\right)
\end{equation}

\noindent
The stable classifier has a constant weight $\omega_s=0.5$ and plays a crucial role in trying to learn the overall trend and direction of concept drift. Conversely, the dynamic classifiers have gradually decaying weights, according to a damped sliding winding approach where each weight is initialized at $\frac{1}{D}$ and then reduced according to its creation time

\begin{equation}
    \omega_d=\left\{\begin{array}{cc}
\omega_d\left(1-\frac{1}{D}\right) & d=1, \ldots, D-1 \\
\frac{1}{D} & d=D
\end{array}\right.
\end{equation}

The most recent classifiers are useful in detecting sudden concept drifts and have highest weights while the old dynamic classifiers have lower weights and can help to identify gradual drifts. The same pairwise strategy based on an ensemble composed by a stable classifier and D dynamic classifiers was used by \cite{Zhang2022}. They modified the original strategy by introducing a reinforcement mechanism to adjust the weights $\omega_d$ according to the prediction performance and the class imbalance issue. The weights adjustment strategy is described by Algorithm \ref{alg:4}. It should be noted that this procedure is only implemented after the true label $y_t$ has been revealed by the oracle. The damped class imbalance ratio (DCIR) value is obtained by taking into account the number of observations for each class collected so far. This is expected to be useful when dealing with imbalanced classes. With regards to the instance selection criterion, the authors consider a hybrid strategy combining uncertainty sampling and random sampling, since approaches solely based on uncertainty could ignore a concept change that is not close to the boundary. \cite{wozniak2023active} recently proposed another ensemble-based active learning strategy where the data points to be labeled are selected from the current chunk using the budget labeling active learning strategy introduced by \cite{zyblewski2020combination}. According to this approach, the learner selects both random and informative data points, where the informativeness is determined using the support function threshold, which in the case of binary classification problems can be interpreted as a distance from the decision boundary.

\begin{algorithm}[h]
\caption{Weight adjustment for dynamic classifiers}\label{alg:4}
\begin{algorithmic}[h]
\Require a labeled observation $(\mathbf{x}_t,y_t)$, number of classes $K$, number of dynamic classifiers $D$, current weights $\omega_d$ with $d=1,\ldots,D$, DCIR for each class $\text{DCIR}{\kappa}$ for $\kappa \in K$.
\If{$\operatorname{DCIR}\left[y_t\right] < \frac{1}{K}$} \Comment{Check if it belongs to the minority class}
\For{$d$ in $(1,D)$}
\If{$C_d(\mathbf{x}_t) = y_t$} \Comment{Check if the prediction made by $C_d$ is correct}
\State $\omega_d \gets \omega_d \left(1 + \frac{1}{D}\right)$ \Comment{Increase weight of classifier $C_d$}
\Else
\State $\omega_d \gets \omega_d \left(1 - \frac{1}{D}\right)$ \Comment{Decrease weight of classifier $C_d$}
\EndIf
\EndFor
\EndIf
\end{algorithmic}
\end{algorithm}

Another way to perform online active learning in time-varying data streams is to use clustering-based approaches. \cite{halder2023autonomic} extended the framework based on stable and dynamic classifiers by introducing a clustering step that aims to train the new stable classifier $C_s$ on the most informative and representative instances from each data block. Similarly, \cite{iencoclustering} investigated a clustering-based approach in a batch-based scenario, where only a fraction of the incoming block of observations can be labeled. They extend the pre-clustering approach \citep{Nguyen2004}, which had been previously studied in the pool-based scenario, to the stream-based case. The sampling strategy takes into account an extra-cluster metric, to sort the clusters, and an intra-cluster one, to sort the observations within each cluster. When a new batch arrives, observations are clustered, and clusters are sorted based on the homogeneity of the clusters, which is measured taking into account the number of (predicted) classes within each cluster. If a cluster is balanced in the number of expected classes, it is regarded to as an uncertain cluster that covers a more difficult area of the input space. Within each cluster, the certainty of an observations is determined by its representativeness, namely the distance from the centroid, and the uncertainty, meant as the maximum a posterior probability among all the predicted classes for $\mathbf{x}_t$. When the clusters and observations are ranked, the learner starts to iteratively ask the observations label in an alternate fashion. To sample the most representative data points from each batch, \cite{zhang2023online} suggested the use of density-peak clustering and recognize the incomplete clusters in the dynamic feature space through the altitude of these data points. This allows to query the observations belonging to those regions in the following iterations.

Recently, \cite{yin2023clustering} proposed an adaptive data stream classification method based on microclustering. After initializing micro-clusters from the initial training data, they collected new labels using a mixed strategy that combines random sampling with a class-weighted margin score. Then, the micro-cluster learning model is dynamically updated to adapt to the presence of concept drifts. 

\begin{figure}[h]
  \centering
  \includegraphics[width=0.7\linewidth]{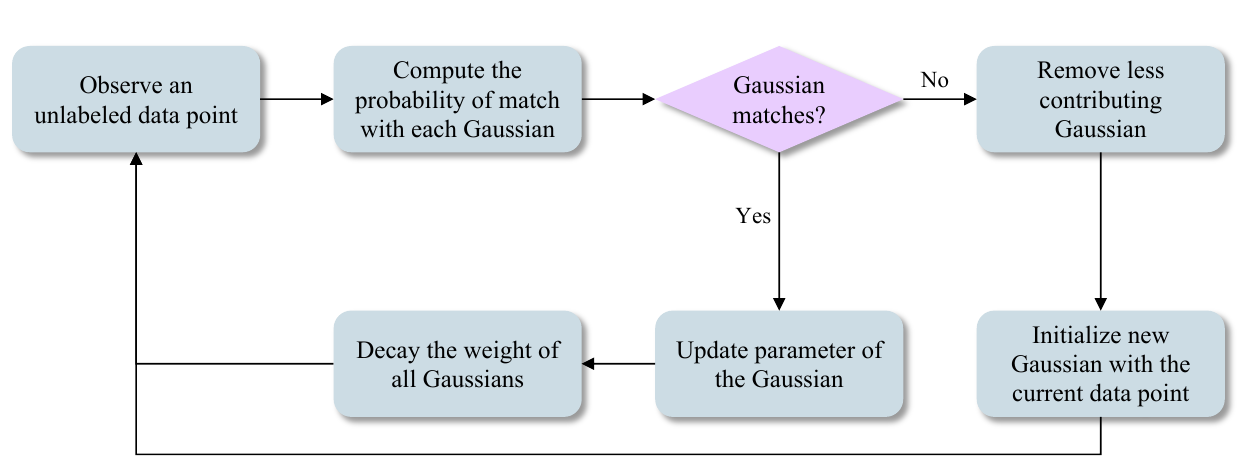}
  \caption{Main steps of the growing Gaussian mixture model used by \cite{Mohamad2018}.}
  \label{fig:gaussian}
\end{figure}

Another approach that tries to exploit the clustering nature of the incoming observations has been proposed by \cite{Mohamad2018}, with the use of bi-criteria active learning algorithm that considers both density in the input space and label uncertainty. The density-based criterion makes use of the growing Gaussian mixture model proposed (GGMM) by \cite{Bouchachia2014}, which is used to find clusters in the data and estimate its density. This model creates a new cluster when a new data point $\mathbf{x}_t$ has a Mahalanobis distance greater than a given closeness threshold from the nearest cluster, among the currently available ones. A flowchart describing the main steps of the GGMM is depicted in Figure \ref{fig:gaussian}.

A Bayesian logistic regression model is used for addressing the label uncertainty criterion and the concept drift. As the classifier parameters $\mathbf{w}_t$ are assumed to evolve over time, the model is incrementally updated using a discrepancy measure, which is computed as the difference between the uncertainty of the model in $\mathbf{x}_t$ before and after the true label $y_t$ is added to the training set. The query strategy follows the b-sampling approach, in trying to sample, with high probability, the observations that contribute the most to the current error. The combination of density and uncertainty is also employed by \cite{Liu2021}, who proposed a cognitive dual query strategy for online active learning in the presence of concept drifts and noise. The local density measure is used to obtain representative instances and the uncertainty criterion aim to select data points where the classifier is less confident. The cognitive aspect takes into account Ebbinghaus’s law of human memory \citep{Ebbinghaus2013} to determine an optimal replacement policy. The proposed strategy tries to tackle both gradual and abrupt drifts. The drift is generally considered as a change in the underlying joint probability distribution from one time step $t$ to another, namely $p_t(\mathbf{x}, y) \neq p_{t+1}(\mathbf{x}, y)$. The local density of an observation $\mathbf{x}_t$ is defined by the number of times that $\mathbf{x}_t$ is the nearest neighbor of other instances \citep{Ienco2014}. Since we are in an online framework, the authors proposed to measure the local density using a sliding window model, referred to as a cognition window. Based on the concept of memory strength, the model determines when the current window is full and needs to be updated. Finally, the labeling decision is taken by using two thresholds, one for the local density and one for the classifier uncertainty.

A different sliding window-based online active learning strategy is the one proposed by \cite{Kurlej2011}. The authors proposed a sliding window approach based on a nearest neighbors classifier. The reference set for the k-nearest neighbors model is a window, and it is updated in two ways: in a first-in-first-out manner or using the examples selected by the active learning strategy. Since the reference set is updated over time, this method can effectively deal with concept drift and time-varying data streams. The sampling strategy is also based on two criteria. The first one is similar to the margin-based approaches, an instance is queried if it has a low distance from two observations belonging to different classes. The second criterion, similar to the greedy sampling strategy, seeks observations that have a large minimum distance from the observations in the current reference set. Both criteria are implemented by setting a threshold on the distances.

A simpler approach for taking into account the time-varying aspect of evolving data stream is to force the model to focus on the most recent observations. Along these lines, \cite{Chu2011} propose a framework based on a Bayesian probit model and a time-decay variant. Online Bayesian learning is used to maintain a posterior distribution of the weight vector of a linear classifier over time $\mathbf{w}_t$, and the time-decay strategies are employed to tackle the concept drift and give more importance to recent observations. They also propose an online approximation technique that can handle weighted examples, which is based upon \cite{Minka2001}. They tested different sampling strategies, built upon an online probit classifier. The instance selection criteria are based on entropy, function-value, and random sampling.

\subsection{Evolving fuzzy systems approaches} \label{subsec:evolving}
\begin{figure}[h]
  \centering
  \includegraphics[width=.8\linewidth]{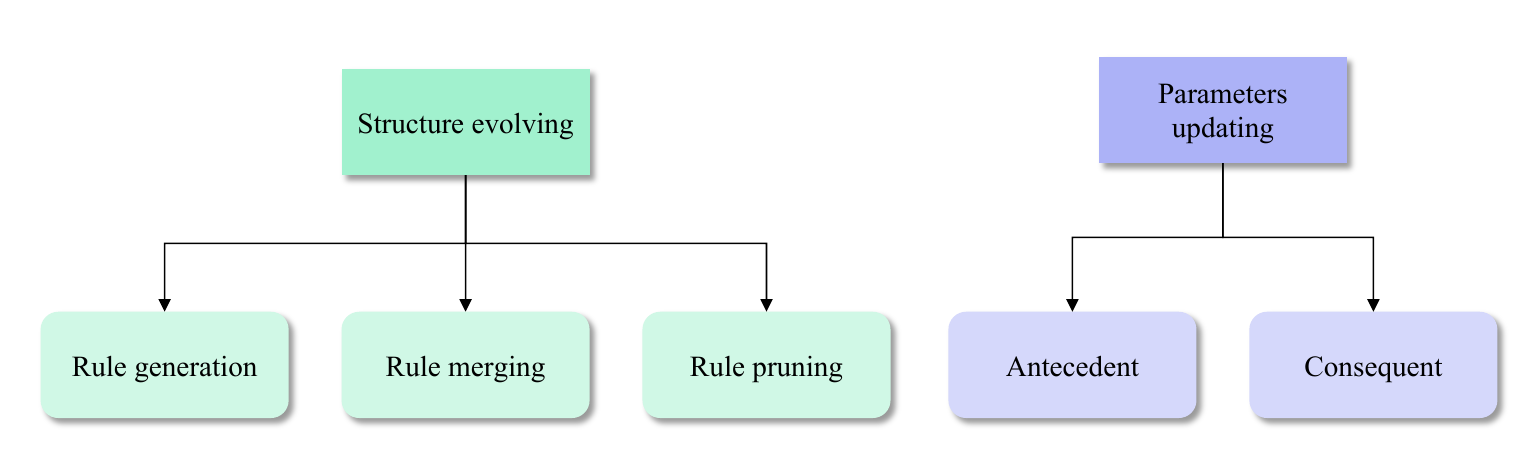}
  \caption{Learning modules of an EFS \citep{Ge2020}.}
  \label{fig:fuzzy}
\end{figure}

An alternative way to take into account the time-varying nature of evolving data streams is the use of evolving fuzzy systems (EFS) \citep{Lughofer2011}, which are soft computing techniques that can efficiently deal with novelty and knowledge expansion. EFS are self-developing, self-learning fuzzy rule-based or neuro-fuzzy systems that self-adapt both their parameters and their structure on-the-fly. They try to mimic human-like reasoning by modeling it with a dynamically developing fuzzy rule-based structure and implementing it utilizing data streams using a formal learning process. The basic rule structure of a fuzzy model is given by

\begin{equation}
\begin{split}
\text{Rule}_i: \textbf{if } & \left(x_1 \textbf{ is } X_{i1}\right) \textbf{ and } \dots \textbf{ and } \left(x_n \textbf{ is } X_{in}\right) \\
& \textbf{ then } \left(y_i=a_{i0}+a_{i1}x_1+\dots+a_{in}x_n\right)
\end{split}
\end{equation}

\noindent
where $\text { Rule }_i \text{ with } i=1,2, \ldots, R$ is one of several fuzzy rules in the current rule base; $x_j(j=1,2, \ldots, n)$ are input variables; $y_i$ denotes the output of the $i$th fuzzy rule; $X_{i j}$ denotes the $j$th prototype (focal point) of the $i$th fuzzy rule; $a_{ij}$ denotes the $j$th parameter of the $i$th fuzzy rule. For a more thorough discussion on EFS and their use in online learning, please see \citep{Lughofer2017,Lughofer2011,Ge2020,Gu2022}. The main components of an EFS are shown in Figure \ref{fig:fuzzy}. The two key components of an EFS are the structure evolving scheme, which contains the rule generation and simplification modules, and the parameters updating scheme. The rule generation module defines when a new rule needs to be added to the current model. The rule merging and pruning steps simplify the models by removing redundant rules and combining two rules when their similarity is larger than a given threshold. The parameter updating modules are used to keep track of the model evolution. These learning modules are used to update the EFS every time a new labeled example $\left(\mathbf{x}_t, y_t\right)$ is made available.

The first single-pass active learning approach based on the use of evolving classification models has been proposed by \cite{Lughofer2012}. The proposed algorithm is based on two key concepts, conflict and ignorance. The former is related to an incoming data point lying close to the boundary between any two classes; the latter considers the distance of the incoming observation from the currently labeled training set, in the feature space. This suggests that the data point falls within a region that has not been thoroughly explored by the learner. Later on, \cite{Lughofer2018} also proposed the first online active learning approach for evolving regression models. Similarly to their previous work \citep{Lughofer2012}, the authors consider the ignorance about the input space in the instance selection criterion. Moreover, they also consider the uncertainty in the model outputs and in the model parameters. The predictive uncertainty is assessed in terms of confidence intervals using locally adaptive error bars. The error bars are inspired by \citep{fuzzyintervals} and the authors propose a new merging approach for dealing with the case of overlapping fuzzy rules. The uncertainty in the model parameters is instead evaluated using the A-optimality criterion, which will be discussed in Section \ref{subsec:bandit} together with other alphabetic optimality criteria. Instead of leveraging the uncertainty about the output, \cite{Pratama2015} set a dynamic threshold based on the variable uncertainty strategy introduced by \cite{Zliobaite2014} while trying to address the what-to-learn question in the training of a recurrent fuzzy classifier. The key idea is that the model is iteratively retrained using data points that fall within rules with low support, which were formed using the smallest amount of observations.  Recently, \cite{lughofer2023online} proposed an online active learning strategy for fuzzy models based on three criteria.
\begin{itemize}
    \item D-optimality in the consequent space to reduce parameter uncertainty, as in \cite{SBAL}.
    \item Overlap degree in the antecedent space to reduce the number of data points lying in the overlap regions of two different rules.
    \item Novelty content in the antecedent space, indicating the required knowledge expansion through rule evolution.
\end{itemize}

A different kind of threshold, based on the spherical potential theory, has been suggested by \cite{Subramanian2014}, with the proposal of a meta-cognitive component that evaluates the novelty content of incoming data points. This is done using a knowledge measure represented by the spherical potential, which has been thoroughly investigated in kernel-based approaches \citep{Hoffmann2007}. The spherical potential is used to set a threshold and decide whether to add a new rule to capture the knowledge in the current sample. It should be noted that the authors also used a threshold based on the prediction error, which could not be used with scarcity of labels. The prediction error is assessed using the hinge loss error function \citep{Suresh2008,Zhang2004}.

Fuzzy models have also been used to solve computer vision tasks. \cite{Weigl2016} analyze the visual inspection quality control case, which is also considered by \cite{rozanec}. They assess the usefulness of the images in a single-pass manner, but the instances that are selected to be queried are accumulated in a buffer, which is later on assigned to an oracle for labeling. Choosing the size of the buffer represents a trade-off problem between timely updating the classifier and requiring continuous interventions from a human annotator. The active learning strategy works by setting a threshold on the certainty of the model with regards to the incoming data points. The authors take into account two model classes, a random forest classifier and an evolving fuzzy classifier. When using random forest, certainty is computed using the best-versus-second-best margin score. Instead, when using evolving fuzzy classifiers, the sample selection criterion takes into account the conflict and ignorance concepts as in \cite{Lughofer2012}.

Finally, \cite{Cernuda2014} combine the use of fuzzy models with a sampling approach inspired by the multivariate statistical process control literature. Indeed, using a latent structure model, they propose a query strategy based on the Hotelling $T^2$ and the squared prediction error (SPE) statistics, which have been extensively used in anomaly detection problems \citep{OAE,Gajjar2018,Vanhatalo2016,Vanhatalo2017}. \cite{Ge2014} used these statistics for pool-based active learning in conjunction with a principal component regression model. The key idea is to use the Hotelling $T^2$ and the SPE statistics to measure the distance between the currently labeled training set and a new unlabeled data point. A high value in one of the two statistics would most likely suggest that the new observation is violating the current model, and thus its inclusion in the training set could bring some valuable information. Similarly, \cite{Cernuda2014} use the Hotelling $T^2$ and the SPE statistics with a partial least squares model. Then, when a new observation is added to the training set, they retrain a TS fuzzy model using a sliding window approach.

\subsection{Experimental design and bandit approaches} \label{subsec:bandit}
Optimal experimental design \citep{Karlin1966} is a research field that is closely related to active learning. It deals with the design of experiments that allow for efficient estimation of model parameters or improved prediction performance while minimizing the number of required labeled examples, also referred to as the number of runs $N$. Many optimality criteria have been developed in thriving to strike a balance between efficient use of resources and ensuring good performance of the model. The traditional framework of optimal experimental designs focuses on linear regression models of the form

\begin{equation}
    \mathbf{y}=\mathbf{X} \boldsymbol{\beta}+\boldsymbol{\varepsilon}
\end{equation}

\noindent
where, given $d$ input variables, $y$ is a $N \times 1$ vector of response variables, $\mathbf{X}$ is a $N \times d$ model matrix, $\boldsymbol{\beta}$ is a $d \times 1$ vector of regression coefficients, and $\boldsymbol{\varepsilon}$ is a $N \times 1$ vector representing the noise, with covariance matrix $\sigma^2 \mathbf{I}$. If the matrix $\mathbf{X}^\top \mathbf{X}$ is of full rank, an ordinary least square (OLS) estimator for $\boldsymbol{\beta}$ can be obtained using

\begin{equation}
    \widehat{\boldsymbol{\beta}}=\left(\mathbf{X}^\top \mathbf{X}\right)^{-1} \mathbf{X}^\top \mathbf{y}
\end{equation}

\noindent
In general, design optimality criteria leverage the information contained in the moment matrix, which is defined as $\mathbf{M}=\mathbf{X}^\top \mathbf{X} / N$. The matrix $\mathbf{X}^\top \mathbf{X}$ plays a crucial role in the estimation of the model coefficients $\boldsymbol{\beta}$, and it is important to perceive information about the design geometry. Indeed, with Gaussian noise characterized by $\boldsymbol{\varepsilon} \sim \mathcal{N}\left(\mathbf{0}, \sigma^2 \mathbf{I}\right)$, we know that

\begin{equation}
    \widehat{\boldsymbol{\beta}} \mid \mathbf{X} \sim \mathcal{N}\left(\boldsymbol{\beta},\left(\mathbf{X}^\top \mathbf{X}\right)^{-1} \sigma^2\right)
\end{equation}

\noindent
and we can define a $100(1-\alpha) \%$ confidence ellipsoid related to the solutions of $\boldsymbol{\beta}$ using

\begin{equation}
    \frac{(\mathbf{b}-\widehat{\boldsymbol{\beta}})^\top\left(\mathbf{X}^\top \mathbf{X}\right)(\mathbf{b}-\widehat{\boldsymbol{\beta}})}{d s^2} \leq F_{\alpha, d, N-d}
\end{equation}

\noindent
where $s^2$ represents the residual mean square, $F_{\alpha, d, N-d}$ is the $100(1-\alpha)$ percentile derived from the Fisher distribution, and $\mathbf{b}$ indicates all the possible vectors that could be the true model parameter $\boldsymbol{\beta}$. The ellipsoid can also be expressed as $(\mathbf{b}-\widehat{\boldsymbol{\beta}})^\top\left(\mathbf{X}^\top \mathbf{X}\right)(\mathbf{b}-\widehat{\boldsymbol{\beta}}) \leq C$, where $C=d s^2 F_{\alpha, d, N-d}$. The volume of this ellipsoid is inversely proportional to the square root of the determinant of $\mathbf{X}^\top \mathbf{X}$, and the length of its axes is proportional to $1 / \lambda_i$, where $\lambda_i$ represents the $i$th eigenvalue of $\mathbf{X}^\top \mathbf{X}$, with $i=1, \ldots, d$. The so-called alphabetic optimality criteria pursuit efficient designs by exploiting these properties \citep{Kiefer1959}. The most commonly employed optimality criteria for good parameter estimation are A-, D- and E-optimality:

\begin{itemize}
    \item \textit{A-optimality. }This criterion pursues good model parameter estimation by minimizing the sum of the variances of the regression coefficients. Knowing that the coefficients variances appear on the diagonal of the matrix $\left(\mathbf{X}^\top \mathbf{X}\right)^{-1}$, it can be shown that an A-optimal design is given by a design $\mathcal{D}^*$ that satisfies $\min _{\mathcal{D}} \operatorname{tr}[\mathbf{M}(\mathcal{D})]^{-1}=\operatorname{tr}\left[\mathbf{M}\left(\mathcal{D}^*\right)\right]^{-1}$.
	\item \textit{D-optimality. }This criterion takes into account both the variance and covariance of the regression coefficients, directly minimizing the total volume of the confidence ellipsoid \citep{Myers2016}. A D-optimal design is given by a design $\mathcal{D}^*$ that satisfies $\max _{\mathcal{D}}|\mathbf{M}(\mathcal{D})|=\left|\mathbf{M}\left(\mathcal{D}^*\right)\right|$ \citep{John1975}.
	\item \textit{E-optimality. }This strategy tries to shrink the ellipsoid by minimizing the maximum eigenvalue of the covariance matrix.
\end{itemize}

\begin{figure}[h]
  \centering
  \includegraphics[width=0.9\linewidth]{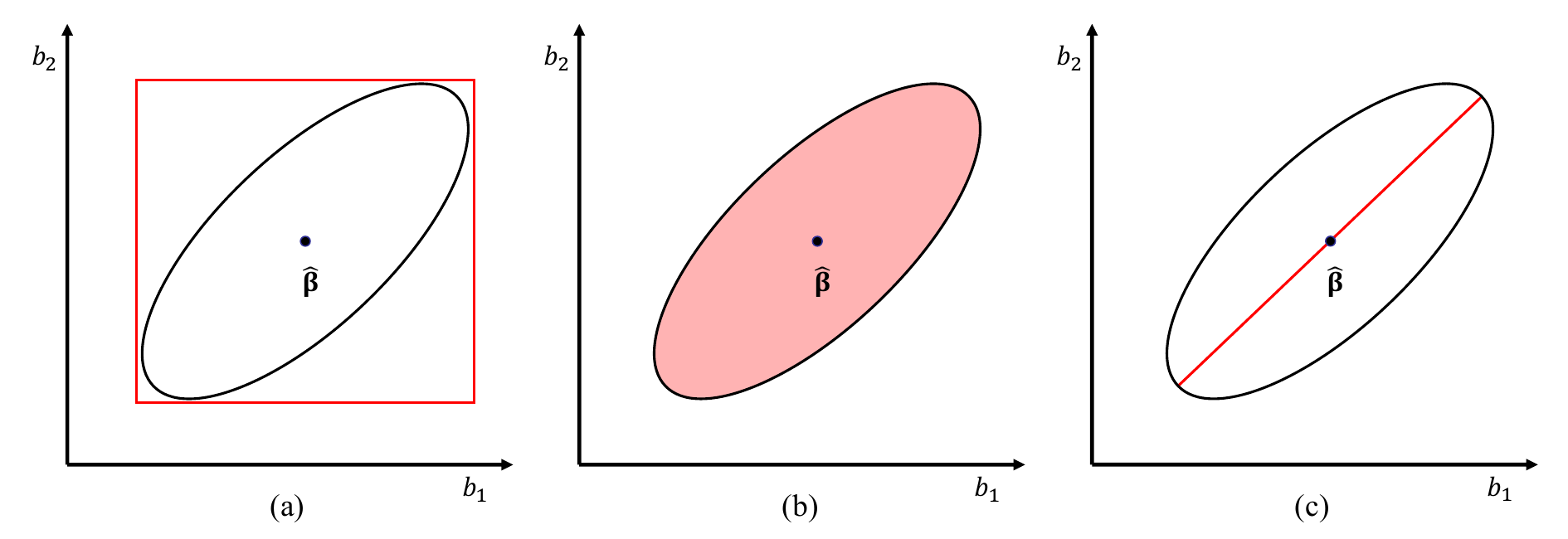}
  \caption{Confidence ellipsoid around the model parameters and optimality criteria: A-optimality (a) shrinks the hyperrectangular enclosing the confidence ellipsoid \citep{Asprey2002,Galvanin2010}, D-optimality (b) aim to shrink the total volume of the ellipsoid, and E-optimality (c) tries to reduce the length of the longest axis \citep{Jamieson2018}.}
  \label{fig:optimality}
\end{figure}

The geometrical intuition behind these criteria is illustrated, in the two-dimensional case, in Figure \ref{fig:optimality}.

Finally, there are also optimality criteria that focus on developing models with good predictive properties. Within this class, \textit{G-optimality} represents a criterion that is used to seek protection against the worst-case prediction variance in a region of interest $\mathcal{R}$. This is achieved by solving

\begin{equation}
    \min _{\mathcal{D}}\left[\max _{\mathbf{x} \in \mathcal{R}} v(\mathbf{x})\right]
\end{equation}

\noindent
where $v(\mathbf{x})$ represents the scaled prediction variance of the current model in the data point $\mathbf{x}$, which can be computed as

\begin{equation}
    v(\mathbf{x})=N \mathbf{x}^{(m) \mathrm{T}}\left(\mathbf{X}^\top \mathbf{X}\right)^{-1} \mathbf{x}^{(m)}
\end{equation}

\noindent
where $\mathbf{x}^{(m)}$ represents the data point where the variance is being estimated, expanded to the model form. It should be noted that G-optimality can be highly influenced by anomalous observations, as it protects against the highest possible variance over all the region $\mathcal{R}$. This issue can be tackled by using \textit{I-} or \textit{V-optimality}, which estimate the overall prediction variance over $\mathcal{R}$ by integrating or averaging, respectively. For a more extensive discussion on optimal designs, please see \cite{Montgomery2012} or \cite{Myers2016}.

The use of optimality criteria has proven to be highly beneficial in offline experimental design, allowing practitioners to pre-determine the location of each design point with ease. However, these methods require modification to be applied in a stream-based scenario where data points arrive sequentially. A common approach for obtaining a near-optimal design with streaming observational data is represented by thresholding. \cite{Riquelme2017} proposed a thresholding algorithm for online active linear regression, which is related to the A-optimality criterion. Their approach uses a norm-thresholding algorithm, where only observations with large, scaled norms are selected. The design is augmented with the observations $\mathbf{x}$ whose norm exceeds a threshold $\Gamma$ given by

\begin{equation}
    \mathds{P}(\|\mathbf{x}\| \geq \Gamma)=\alpha
\end{equation}

\noindent
where $\alpha$ is the ratio of observations we are willing to label out of the incoming data stream. Another approach related to the A-optimality criterion was proposed by \cite{Fontaine2021}, who studied online optimal design under heteroskedasticity assumptions, with the objective of optimally allocating the total labeling budget between covariates in order to balance the variance of each estimated coefficient. \cite{SBAL} further extended the thresholding approach introduced by \cite{Riquelme2017} by proposing a conditional D-optimality (CDO) algorithm. The terms conditional refers to the fact the design is marginally optimal, given an initial set of labeled observations to be augmented. The main steps of the CDO approach are reported in Algorithm \ref{alg:5}. The authors exploited the connection between D-optimality and prediction variance previously highlighted by \cite{Myers2016}. The sampling strategy selects observations by setting a threshold $\Gamma$ given by

\begin{equation}
    \mathds{P}\left(\mathbf{x}_t^\top\left(\mathbf{X}^\top \mathbf{X}\right)^{-1} \mathbf{x}_t \geq \Gamma\right)=\alpha
\label{eq:cdo}
\end{equation}

\noindent
where $\mathbf{X}$ is the current set of labeled observations and $\mathbf{x}_t$ is the data point that is currently under evaluation. The threshold is estimated using kernel density estimation (KDE) on a set of  $j$ unlabeled observations, which are taken passively from the data stream without querying any label. This provides an initial set of data, referred to as warm-up set, that can be used to estimate the covariance matrix and the threshold.

\begin{algorithm}[h]
\caption{Online active learning using CDO}\label{alg:5}
\begin{algorithmic}[h]
\Require an initial random design $\mathbf{X}$, a data stream $\mathbf{S}$, a warm-up length $j$, a sampling rate $\alpha$, a budget $B$
\State $t \gets 1$ \Comment{Timestamp}
\State $c \gets 0$ \Comment{Labeling cost}
\State Set $\mathbf{W} = \varnothing$ \Comment{Warm-up set to estimate $\boldsymbol{\Sigma}$ and $\Gamma$}
\While{$t \leq j$}
    \State Observe incoming data point $\mathbf{x}_t \in \mathbf{S}$
    \State Select $\mathbf{x}_t: \mathbf{W}=\mathbf{W} \cup \mathbf{x}_t$
    \State $t \gets t + 1$
\EndWhile
\State Estimate the covariance matrix $\mathbf{\Sigma}$ of $\mathbf{W} $ and perform eigendecomposition $\boldsymbol{\boldsymbol{\Sigma}} = \mathbf{U}\mathbf{\Lambda}\mathbf{U}^\top$
\State Whiten the initial design by computing $\mathbf{Z} = \Lambda^{-1/2} \mathbf{U}^\top \mathbf{X}$
\State Whiten the warm-up observations by computing $\mathbf{V} = \mathbf{\Lambda}^{-1/2} \mathbf{U}^\top \mathbf{W} $
\State Estimate $\Gamma$ using KDE on $\mathbf{V}$ with the desired sampling rate $\alpha$ using Equation \ref{eq:cdo} with $Z$ and $V$
\While{$c \leq B$ and $t \leq |\mathbf{S}|$}
    \State Observe incoming data point $\mathbf{x}_t \in \mathbf{S}$
    \State Whiten $\mathbf{x}_t$ by computing $\mathbf{z}_t = \mathbf{\Lambda}^{-1/2} \mathbf{U}^\top \mathbf{x}_t$
    \If{$\mathbf{z}_t^\top (\mathbf{Z}^\top \mathbf{Z})^{-1} \mathbf{z}_t \geq \Gamma$}
        \State Ask for the label $y_i$ and augment the labeled dataset: $\mathbf{Z} \gets \mathbf{Z} \cup \mathbf{z}_t$
        \State $c \gets c+1$ \Comment{Pay for the label}
        \State Update threshold $\Gamma$ to measure the prediction variance of the enlarged design
    \Else
        \State Discard $\mathbf{x}_t$
    \EndIf
\State $t \gets t+1$
\EndWhile
\end{algorithmic}
\end{algorithm}

\noindent
\cite{ROAL} also investigated how the presence of outliers affect the performance of online active linear regression strategies. They showed how the design optimality-based sampling strategies might be attracted to outliers, whose inclusion in the design eventually degrades the predictive performance of the model. This issue can be tackled by bounding the search area of the learner with two thresholds, as in

\begin{equation}
    \mathds{P}\left(\Gamma_1 \leq \mathbf{x}_t^\top\left(\mathbf{X}^\top \mathbf{X}\right)^{-1} \mathbf{x}_t \leq \Gamma_2\right)=\alpha
\end{equation}

\noindent
where the choice of $\Gamma_2$ represents a trade-off between seeking protection against outliers and exploring uncertain regions of the input space. 

The norm-thresholding approach was also extended by \cite{Riquelme2017multi} to the case where the learner tries to estimate uniformly well a set of models, given a shared budget. This scenario is similar to a multi-armed bandit (MAB) problem where the learner wants to estimate the mean of a finite set of arms by setting a budget on the number of allowed pulls \citep{Ruan2020,Audibert2010,Jamieson2014,Soare2013}. The authors propose a trace upper confidence bound (UCB) algorithm to simultaneously estimate the difficulty of each model and allocate the shared labeling budget proportionally to these estimates. UCB is a common algorithm used in MAB problems to balance exploration and exploitation \citep{Carpentier2015,Garivier2008}, which takes into account the predicted mean value and the predicted standard deviation, weighted by an adjustable parameter \citep{Thompson2022}. This allows to balance the exploitation of data points with a high predicted value and the exploration of areas with high uncertainty. 

In general, MAB problems can be seen as a special case of sequential experimental design, where the goal is to sequentially choose experiments to perform with the aim of maximizing some outcome. The typical framework of a MAB problem can be regarded as an optimization problem where the learner must identify the option or arm with the highest reward, among a set of available arms characterized by different reward distributions. Both MAB and active learning paradigms involve a sequential decision-making process where the learner aims to maximize a reward or improve model accuracy by selecting an arm to pull or a data point to label, respectively, and receiving feedback (in the form of a reward or label request) for each selection. There are two main approaches to tackle MAB problems:

\begin{itemize}
    \item \textit{Regret minimization. }This approach is coherent with the objective of maximizing the cumulative reward observed over many trials. In this case, the learner must balance exploration, namely trying out different arms to learn more about the reward distributions, with exploitation, i.e., using current knowledge to choose the most promising arm. These kinds of algorithms strike a balance between learning a good model and obtaining high rewards. A few examples might be treatment design, online advertising and recommender systems.
    \item \textit{Pure exploration. }In this case, we are interested in finding the most promising arm, with a certain confidence or given a fixed budget on the number of pulls. To do so, the objective is to learn a good model while minimizing the number of measurements or labels required. This scenario is suggested in circumstances where, due to safety constraints, we are not given complete freedom to change the variable levels and we are mostly interested in understanding the underlying model governing the system. Possible examples include drug discovery or soft sensor development \citep{Fortuna2007,Shi2018,Chan2018,Tang2018}.
\end{itemize}

\noindent
The pure exploration approach is particularly useful when coupled with the study of linear bandits, which are a type of contextual bandit algorithms that assume a linear relationship between the features of the context and the expected reward of each arm. In this type of problem, when an arm $\mathrm{x} \in \mathcal{X}$ is pulled, the learner observes a reward $r(\mathbf{x})$ that depends on an unknown parameter $\boldsymbol{\theta}^* \in \mathds{R}^d$ according to the linear model

\begin{equation}
    r(\mathbf{x})=\mathbf{x}^\top \boldsymbol{\theta}^*+\varepsilon
\end{equation}

\noindent
where $\varepsilon$ is a zero-mean i.i.d. noise. This is similar to active linear regression in that, in both cases, the learner aims to select the most informative data points to learn about the underlying model or system \citep{Audibert2010,Jamieson2014}. \cite{Soare2014}, investigated this problem, in the offline setting, using the G-optimality criterion and a newly proposed $\mathcal{XY}$-allocation algorithm. \cite{Jedra2020} proposed a fixed-confidence algorithm for the same problem, while \cite{Azizi2022} analyzed the fixed-budget case, extending the framework to the case where the underlying model is represented by a generalized linear model \citep{Filippi2010}. An interesting variant of this problem is presented in the study of transductive experimental designs. A transductive design is a problem where we can pull arms from a set $\mathcal{X} \in \mathds{R}^d$, with the objective of identifying the best arm or improve the predictions over a separate set of observations $\mathcal{Z} \in \mathds{R}^d$, which is given, in an unlabeled form, beforehand. A practical example of this case is when we are trying to infer the user preferences over a set of products, but we can only do that by pulling arms from a limited set of free trials. Alternatively, we might be interested in estimating the efficacy of a drug over a certain population, while doing experiments on a population with different characteristics. This problem has been tackled with an active learning approach by \cite{Yu2006}, with the idea of exploiting unlabeled data points in $\mathcal{Z}$ while evaluating the informativeness of the data points in $\mathcal{X}$. The transductive case of sequential experimental design has been explored by \cite{Fiez2019}, but instead of performing active learning, they were interested in inferring the best reward over $\mathcal{Z}$, only pulling the arms in $\mathcal{X}$. Finally, this has been extended to the online scenario by \cite{Camilleri2021}, balancing the trade-off between time complexity and label complexity, namely between the number of unlabeled observations spanned and the number of labels queried in order to stop the learning procedure and declare the best-arm.

In addition to MAB, reinforcement learning-based approaches can also be applied to active learning in order to optimize a decision-making policy that balances the exploration of uncertain data with the exploitation of information learned from previous observations. This can be particularly useful in applications where the goal is to maximize the expected cumulative reward over time, such as in robotics or game playing. Compared to MAB, reinforcement learning-based approaches offer a more general and flexible framework for active learning, allowing for a wider range of problem formulations and feedback signals \citep{Menard2021,Fang2017,Rudovic2019}. One approach to combining active learning and reinforcement learning is through modeling the sampling routine as a contextual-bandit problem, as proposed by \cite{RALWasserman}. In this approach, the rewards are based on the usefulness of the query behavior of the learner. The key intuition behind the use of reinforcement learning in online active learning is that the learner gets feedback after the requested label, based on how useful the request actually was. In contrast to the traditional active learning view, where most of the effort is dedicated to the instance selection phase, the learner is penalized ex-post for querying useless instances. The learner gets a positive reward $\rho^{+}$ if it asks for the label when it would have otherwise predicted the wrong class, and a negative reward $\rho^{-}$ when querying was unnecessary as the model would have predicted the right label. The contextual bandit problem is implemented by building an ensemble of different models, with each expert suggesting whether to query or not based on whether its prediction certainty exceeds a threshold $\Gamma$. The models are assigned a decision power based on how past suggestions were rewarded and how coherent they were with the other experts' suggestions. When an observation is sent to the oracle for labeling, the reward is computed, and the objective function of the learner is to maximize the total reward over a time horizon $T$. 

Another reinforcement learning-based approach has been proposed by \cite{Woodward2017}. They considered the case where at each time step $t$ the learner needs to decide whether to predict the label of the unlabeled data point $\mathbf{x}_t$ or pay to request its label $y_t$. The reinforcement learning framework is used to find an optimal policy $\pi^*\left(s_t\right)$ that takes into account the cost of asking for a label and the cost of making an incorrect prediction, where $s_t$ represents the state that is given in input at the time$ t$ to a policy $\pi\left(s_t\right)$ that outputs the suggested action $a_t$. The authors approximate the action-value function using a long short-term memory (LSTM) neural network with a linear output layer. The optimal policy is determined by maximizing the long-term reward, after assigning a reward to a label request $R_{req}$, a correct prediction $R_{corr}$, and an incorrect prediction $R_{inc}$. It should be noted that $R_{corr}$ and $R_{inc}$ should be negative rewards, as they are associated with costly actions.

\section{Evaluation strategies} \label{sec:evaluation}
The use of active learning approaches is becoming increasingly common in machine learning, allowing models to be trained more efficiently by selecting the most informative examples for labeling. To evaluate the performance of these approaches, it is typical to compare them to a passive random sampling strategy by generating learning curves that plot the model performance (e.g., accuracy, F1 score, or root mean square error) on a holdout test set over the number of labeled examples used for training. Learning curves are a useful tool for comparing the asymptotic performance of different strategies and their sample efficiency, with the slope of the curve reflecting the rate at which the model performance improves with additional labeled examples. A steeper slope indicates a more sample-efficient strategy. When multiple sampling strategies are being compared, a visual inspection of the learning curves may not be sufficient, and more rigorous statistical tests may be necessary. \cite{Reyes2018} recommend the use of non-parametric statistical tests to analyze the effectiveness of active learning strategies for classification tasks. The sign test \citep{Steel1959} or the Wilkinson signed-ranks test \citep{Wilcoxon1945} can be used to compare two strategies, while the Friedman test \citep{Friedman1940}, the Friedman aligned-ranks test \citep{Hodges1962}, the Friedman test with Iman-Davenport correction \citep{Iman1980}, or the Quade test \citep{Quade1979} can be used when evaluating more than two strategies. These statistical tests can provide insight into whether the difference in performance between the active learning and passive random sampling strategies is statistically significant. 

\begin{algorithm}[h]
\caption{Prequential evaluation for online active learning}\label{alg:6}
\begin{algorithmic}[h]
\Require an initial model $\mathbf{w}_0$, a data stream $\mathbf{S}$, a budget $B$, an active learning strategy $Q$.
\State $t \gets 1$ \Comment{Timestamp}
\State $\mathbf{P} \gets \emptyset$ \Comment{Storing predictions}
\While{$c \leq B$ and $i \leq |\mathbf{S}|$}
    \State Observe the data point $\mathbf{x}_t \in \mathbf{S}$
    \State Predict the label $\widehat{y}_t$ and store it: $\mathbf{P} \gets \mathbf{P} \cup {\widehat{y}_t}$
    \If{$Q(\mathbf{x}_t) = \texttt{True}$} \Comment{Sampling decision}
        \State Ask for the true label $y_t$ and update the model
        \State $c \gets c+1$ \Comment{Pay for the label}
    \Else
        \State Discard $\mathbf{x}_t$
    \EndIf
\State $t \gets t+1$
\EndWhile
\end{algorithmic}
\end{algorithm}

Overall, the use of learning curves and statistical tests can provide valuable insights into the effectiveness and efficiency of different active learning strategies. By understanding the statistical significance of differences in performance between these strategies, researchers can make informed decisions about which approaches are more effective for a particular task or dataset. Furthermore, the choice of the evaluation scheme is crucial when assessing the performance of active learning approaches. If we use an evaluation scheme based on a holdout test set, at each learning step $t$ the performance of the model is assessed using the same test set. This can be a reasonable approach if we are dealing with a stationary data stream, which does not evolve over time. Under these assumptions, using the same test set we might be able to better assess the prediction improvement as more labeled examples are included in the design. However, this approach might not be ideal when dealing with drifting data streams. In these circumstances, a prequential evaluation scheme can be more useful to monitor the evolution of the prediction error over time \citep{covid,Cerqueira2020,Tieppo2022,Cacciarelli2021}. In online learning, prequential evaluation is also referred to as test-then-train approach, and it involves using each incoming instance first to measure the prediction error, and then to be included in the training set \citep{quintana}. The main steps of the test-then-train approach are reported in Algorithm \ref{alg:6}. The key idea is that at each time step $t$, we first test the model by making a prediction, then we decide whether to query the true labels and finally we update our model. 

An in-depth analysis and discussion between the use of a holdout test set and the prequential evaluation scheme for streaming data has been provided by \cite{gama2009issues,gama2013evaluating}, who suggested the use of a prequential evaluation scheme with forgetting mechanisms. For scenarios with imbalanced data streams, a specialized prequential variant of the area under the curve metric has been proposed by \cite{Brzezinski2015,Brzezinski2017}. From an implementation perspective, \cite{bifet2010moa} developed an open source software suite called MOA for data stream mining, which includes both the holdout and prequential strategies. This framework has found widespread application in the evaluation of online active learning strategies, as evidenced by the studies conducted by \cite{Liu2021,Shan2019,Weigl2016,Zhang2020,alabdulrahman2016active}.

\begin{table}[h]
\centering
\begin{tabular}{|c|p{8cm}|} 
\hline
\textbf{Evaluation Strategy} & \textbf{Works} \\
\hline
Holdout test set & \cite{Desalvo2021,RALWasserman,rozanec,Narr2016,Ferdowsi2013,Bordes2005,Suzuki2021,Ghassemi2016,Qin2021,Woodward2017,Riquelme2017AAAI,SBAL,ROAL,Manjah} \\
\hline
Prequential/Test-then-train & \cite{Zhang2022,Pham2022,Castellani2022,Chu2011,Zhang2018,Krawczyk2018,Xu2016,Mohamad2020,Weigl2016,iencoclustering,Zhang2020} \\
\hline
\end{tabular}
\caption{Evaluation strategies.}
\label{tab:evaluation}
\end{table}

In Table \ref{tab:evaluation}, we categorize the studies based on the experimental protocols they employed to evaluate the sampling strategies. The table exclusively includes approaches where the evaluation strategy was explicitly defined. In most cases, when assessing active learning methods in the context of drifting data streams, a prequential approach is favored. Conversely, for scenarios where the methods are ill-suited to handle concept drifts, holdout test sets tend to be the preferred choice. In approaches not featured in the table, the evaluation strategies exhibited some variations or lacked explicit specification. For instance, in the work by \cite{fujikashima}, their evaluation strategy involved training models on the queried data and subsequently testing them with the entire dataset. This approach differs from the conventional test-then-train paradigm since, in this case, models are tested on data they encountered during training, at least in part. Another example is found in \cite{Zhu2007}, who utilized a window-based approach, assessing prediction accuracy across all observations in the current batch. On a different note, \cite{Hao2018Expert} employed the per-round regret metric, which quantifies the loss difference between the forecaster and the best expert at each iteration of the active learning process. In some instances, none of the previously mentioned methods were employed, as the analysis took a more theoretical perspective. This is exemplified by the works of \cite{Dasgupta2005,Chae2021,Huang2022}. Lastly, bandit algorithms employed a distinct evaluation approach, often aiming to identify the most promising arm with a fixed confidence or budget. In the fixed confidence setting, performance typically hinges on comparing label complexity to problem dimensionality or the number of arms pulled, as observed in \cite{Fiez2019}. Alternatively, regret or error metrics were evaluated against the required number of trials, as demonstrated in the studies by \cite{Riquelme2017multi,sudarsanam2018using,Fontaine2021}.

\section{Real-world applications and challenges} \label{sec:application}
\subsection{Applications}
\begin{figure}[h]
  \centering
  \includegraphics[width=0.9\linewidth]{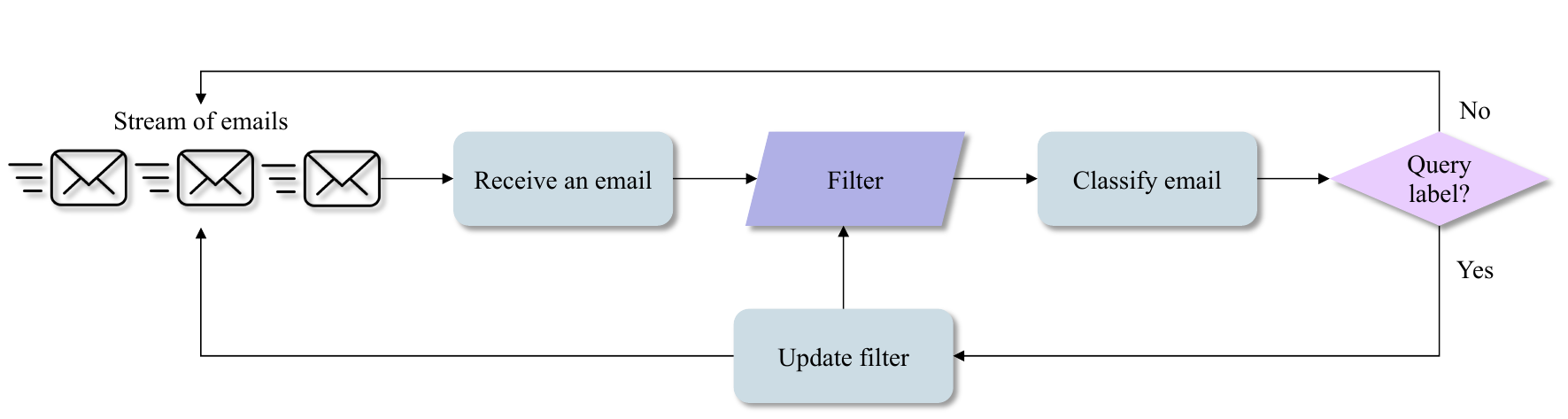}
  \caption{Low-cost active spam filtering \citep{Sculley2007}.}
  \label{fig:spam}
\end{figure}

Online active learning has been recognized as a powerful technique in scenarios where data is arriving at a high velocity, labeling data is expensive, and it is infeasible to store all the unlabeled data before making a decision about which observations to query to update the model. In particular, these techniques have proven particularly useful in dynamic and ever-evolving environments, where models need to adapt to new data in real-time, by selectively querying the most informative instances. One of the first real-world applications of online active learning has been presented by \cite{Sculley2007}, who investigated the scenario of low-cost active spam filtering (Figure \ref{fig:spam}) where a filter is updated online by selecting the most informative emails in real time. Another application of online active learning in the field of IT has been recently presented by \cite{Zhang2020}. They analyzed the scenario of network protocol identification and proposed a method (presented in Section \ref{subsec:drifting}) to select the most representative instances on the fly and adapt the model to dynamic data distributions.

Computer vision is another interesting area where online active learning can be applied. Deep learning models require a large amount of annotated data, making manual annotation of thousands of images one of the most challenging aspects of model development. However, it is important to note that the most effective deep active learning methods proposed so far are not easily adaptable to a stream-based setting. Many of these methods involve clustering or measuring pairwise similarity among image embeddings \citep{Sener2017,Agarwal2020,Ash2019,Citovsky2021,Prabhu}, which cannot be easily done in a single-pass manner. As a result, most online applications of active learning in computer vision rely on the use of traditional models with uncertainty-based sampling. \cite{Narr2016} analyze the stream-based active learning problem for the classification of 3D objects. They used a mondrian forest classifier \citep{Lakshminarayanan2014}, which is an efficient alternative of random forest for the online learning scenario, and selected images with high classification uncertainty to be labeled. \cite{rozanec} used online active learning to reduce the data labeling effort while performing vision-based process monitoring. Initially, features are extracted from the images using a pre-trained ResNet-18 model \citep{He2015} and then, using the mutual information criterion \citep{Kraskov2004}, only $\sqrt{n}$ features \citep{Hua2005} are retained to fit an online classifier, where $n$ is the total number of observations in the training set. The authors combine a simple active learning strategy based on model uncertainty with five streaming classification algorithms, including Hoeffding tree \citep{Hulten2001}, Hoeffding adaptive tree \citep{Bifet2009}, stochastic gradient tree \citep{Gouk2019}, streaming logistic regression, and streaming k-nearest neighbors. Recently, \cite{pmlr-v202-saran23a} proposed a novel approach to streaming active learning with deep neural networks. Given a neural network with $f$ with parameters $\theta$, last-layer parameters $\theta_L$, and the cross-entropy function $\ell$, they compute the gradient representation of the data point $\mathbf{x}_t$, which is given by
\begin{equation}
    g(\mathbf{x}_t) = \frac{\partial}{\partial \theta_L}\ell\left(f(\mathbf{x}_t; \theta),\widehat{y}_t\right)
\end{equation}

\noindent
where $\widehat{y}_t=\operatorname{argmax}f(\mathbf{x}_t; \theta)$. Then, the data points to be included in the batch for training the model are chosen by using a probability $p_t$ proportional to the contribution of the current example to the covariance matrix of the examples collected so far, as in
\begin{equation}
    p_t \propto \operatorname{det}\left(\widehat{\Sigma}_t + g(\mathbf{x}_t)g(\mathbf{x}_t)^\top\right)
\end{equation}

\noindent
where $\widehat{\Sigma}_t$ is the covariance matrix of the data points that have been selected to be included int he current batch, up to the time step $t$.

\begin{figure}[h]
  \centering
  \includegraphics[width=0.6\linewidth]{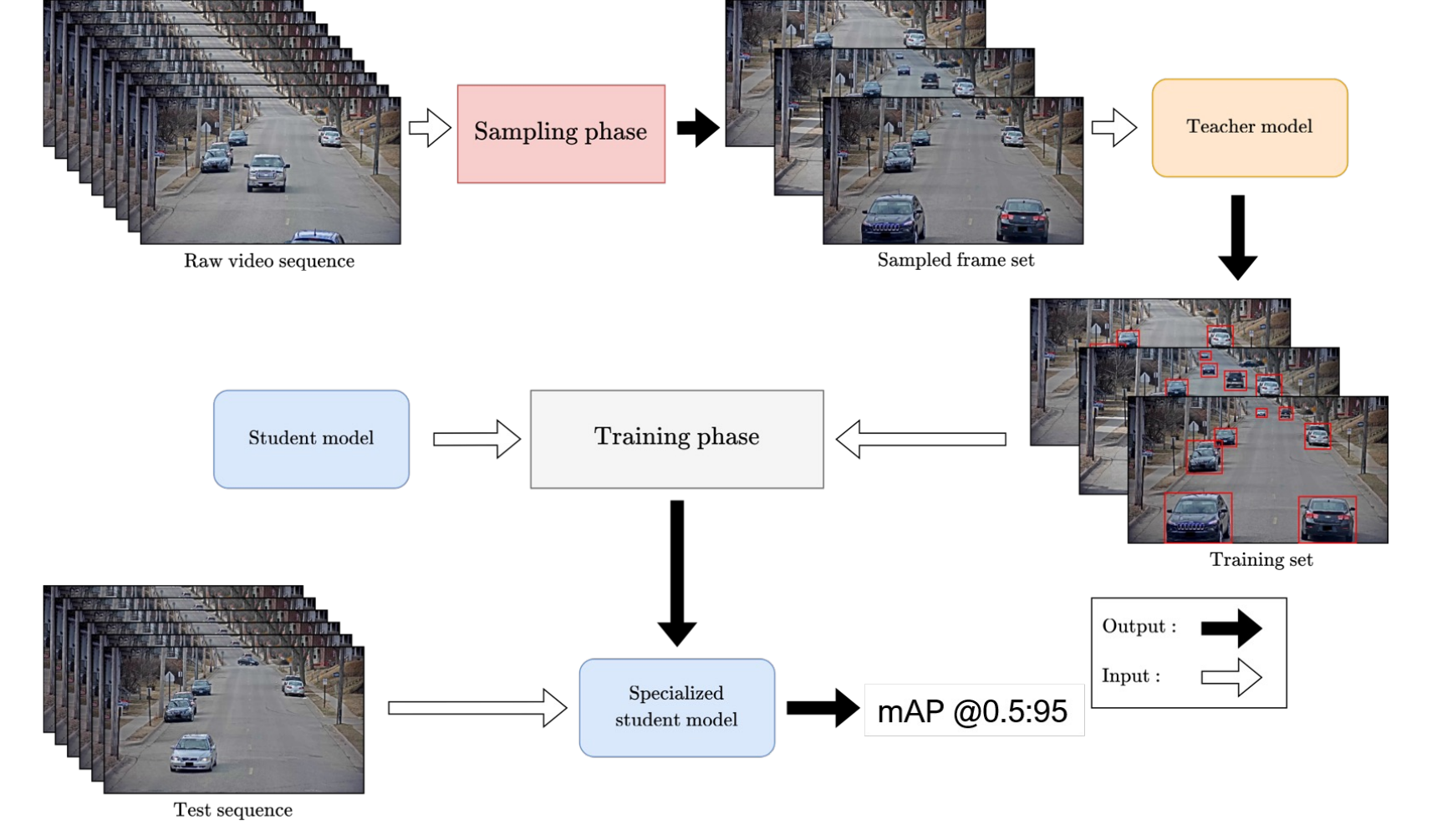}
  \caption{SBAD framework \citep{Manjah}: sampling, fine-tuning and evaluation. The sampling is performed in a single-pass manner via thresholding.}
  \label{fig:sbad}
\end{figure}

Online active learning has also been explored for object detection tasks. \cite{Manjah} proposed a stream-based active distillation (SBAD) framework by combining the concepts of active learning and self-supervision as described in Section \ref{subsec:ssl}. The SBAD framework enables the deployment of scalable deep-learning models as it does not rely on human annotators and takes into account the imperfection of the oracle when distilling knowledge from a large teacher model to a lightweight student. Indeed, the authors suggest setting a threshold on the confidence of the images and only querying images with high confidence in trying to avoid confirmation bias. The threshold is determined using a warm-up phase, similarly to the approach proposed by \cite{SBAL} presented in Algorithm \ref{alg:5}. The SBAD pipeline for model development and evaluation is reported in Figure \ref{fig:sbad}.

The problem of performing active learning for object detection with streaming data has also been explored by \cite{beck2023streamline}. In the case of a camera placed on an autonomous vehicle, the collected data encompasses various scenarios, including clear weather, foggy conditions, and rainy weather, all of which require the model to perform effectively. However, the frequency of these scenarios can vary significantly. In situations where one scenario is prevalent, a passive sampling strategy could tend to sample very few examples from the most rare slices. Instead, the proposed streamline approach by attempts to smartly allocate the budget to obtain more observations from the slices where the model is under-performing. The case of autonomous cars was also considered by \cite{yan2023online}, who used a diversity-based online active learning strategy to reduce false alarm rate and learn unseen faults.

Another interesting industrial application has been recently presented by \cite{ghiasi2023combining}. They proposed a deployable framework that combines a thermodynamics-based compressor model and a Gaussian Process-based surrogate model with an online active learning module. The objective of the study was to minimize the power absorbed by the machine during the boil off process of centrifugal compressor. In the proposed framework, the simulator, the surrogate model, and the optimizer interact in real time to determine the new experimental points.

\subsection{Challenges}
When applying online active learning strategies to real-world problems, there are several potential issues to consider, including:

\begin{itemize}
    \item \textit{Algorithm scalability. }Online active learning algorithms need to be efficient and scalable to handle large datasets and high-velocity data streams. As the amount of data grows, the computational demands of active learning can become prohibitive, making it difficult to deploy in practice. The time required to make the sampling decision needs to be lower than the feed rate of the process being analyzed. If the algorithm is too slow, it may require a buffer, which reduces the benefits of online active learning.
    \item \textit{Labeling quality. }Most online active learning strategies rely heavily on the quality of labeled data, which can be challenging to ensure in real-world scenarios. Human annotators may make errors, introduce biases, or interpret labeling instructions differently. For this reason, in real-life situations, it may be necessary to consider oracle imperfections like in the knowledge distillation case \citep{Baykal2022}. Another difficult aspect related to labeling quality is the delay or latency, which has been described in Section \ref{subsubsec:oal}.
    \item \textit{Data drift. }In real-world settings, data distributions may shift over time, making it challenging for models to adapt and continue providing accurate predictions. Changes in the data distribution may also affect the quality of the labeled data, as the criteria for selecting informative instances may become less effective. Methods from Sections \ref{subsec:drifting} and \ref{subsec:evolving} should be used when dynamic and ever-changing behaviors are expected.
    \item \textit{Model interpretability. }Besides simply asking for the most informative instances from a modeling perspective, it might be useful to provide additional information on why a particular instance is beneficial for improving the performance of the current model. In fields like healthcare and manufacturing this might help practitioners to improve their understanding of the underlying problem.
    \item \textit{Evaluation. }When developing active learning methods from a research perspective, the different query strategies are evaluated assuming the ground-truth labels to be available for a held-out test set, or for the data stream being analyzed. However, in real life, the key motivation behind active learning is label scarcity and thus it might be difficult to thoroughly assess the effectiveness of the deployed sampling strategy.
    \item \textit{Human-computer interaction. }In the context of active learning for data streams, the synergy between human labelers and computer systems plays a pivotal role in the labeling process. While the majority of online active learning methods focus on querying the most informative data points in real-time, we can distinguish between two distinct labeling scenarios:
    \begin{enumerate}
        \item \textit{Real-time annotation. } In most of the presented works, it is assumed that labels are immediately available when a data point is queried from the stream. This immediate access to true labels enables an optimized active learning routine, as the model can be promptly updated and can recommend exploration of new regions based on up-to-date information. However, this approach poses some implementation challenges that need to be addressed with the use of advanced data annotation tools \citep{feuz2013real}.
        \item \textit{Postponed annotation. } There are cases where we must allow for a delay between data querying and labeling. For instance, methods that consider verification latency \citep{Castellani2022,Pham2022} take into account the possibility of delayed labels. This is particularly relevant in situations where a physical quality inspection or medical treatment must occur before the label is revealed. Another example is in the training of deep neural networks, where real-time sampling from a data stream is necessary due to memory constraints \citep{Manjah}, but the labeling and model update phase may occur when a batch is collected, following a batch-mode active learning strategy \citep{Ren2022}.
    \end{enumerate}
\end{itemize}

\section{Summary and future directions} \label{sec:summary}
This survey outlines the challenge of conducting active learning with data streams and investigates different approaches for selecting the most informative data points in real-time. 

\begin{table}[]
\begin{tabular}{@{}|p{2.5cm}|p{1.5cm}|p{1.6cm}|p{1.5cm}|p{4cm}|@{}}
\toprule
\textbf{Data processing} & \textbf{Data stream} & \textbf{Task} & \textbf{Model} & \textbf{Work(s)} \\ \midrule
\multirow{8}{*}{Single-pass} & \multirow{4}{*}{Stationary} & \multirow{2}{*}{Classification} & Single Model & \cite{cesabianchi1,cesabianchi2,Dasgupta2005,Sculley2007,Lu2016,Hao2018,Ghassemi2016,Shah2020,Mohamad2020,pmlr-v202-saran23a,rozanec,Woodward2017} \\ \cmidrule(l){4-5} 
 &  &  & Ensemble & \cite{Huang2022,Desalvo2021,Loy2012,Hao2018Expert,Chae2021} \\ \cmidrule(l){3-5} 
 &  & Regression & Single Model & \cite{Riquelme2017,Fontaine2021,SBAL,ROAL,Cacciarelli2022} \\ \cmidrule(l){3-5} 
 &  & Object detection & Single Model & \cite{Manjah} \\ \cmidrule(l){2-5} 
 & \multirow{2}{*}{Drifting} & \multirow{2}{*}{Classification} & Single Model & \cite{Krawczyk2018,Castellani2022,Pham2022,yin2023clustering,Mohamad2018,Liu2021,Kurlej2011,Chu2011} \\ \cmidrule(l){4-5} 
 &  &  & Ensemble & \cite{Zhang2020,Shan2019,Zhang2018,Zhang2022} \\ \cmidrule(l){2-5} 
 & \multirow{2}{*}{Evolving} & Classification & Single Model & \cite{Lughofer2012,Pratama2015} \\ \cmidrule(l){3-5} 
 &  & Regression & Single Model & \cite{Lughofer2018,lughofer2023online} \\ \midrule
\multirow{5}{*}{Batch} & \multirow{2}{*}{Stationary} & Classification & Single Model & \cite{Bordes2005,Qin2021,fujikashima} \\ \cmidrule(l){3-5} 
 &  & Object detection & Single Model & \cite{beck2023streamline} \\ \cmidrule(l){2-5} 
 & \multirow{2}{*}{Drifting} & \multirow{2}{*}{Classification} & Single Model & \cite{cheng2023active,martins2023meta,iencoclustering,zhang2023online,yan2023online} \\ \cmidrule(l){4-5} 
 &  &  & Ensemble & \cite{Zhu2007,wozniak2023active,halder2023autonomic} \\ \cmidrule(l){2-5} 
 & Evolving & Classification & Single Model & \cite{Subramanian2014,Weigl2016,Cernuda2014} \\ \bottomrule
\end{tabular}
\caption{Online active learning strategies: summary based on data processing capabilities, assumptions about the data stream, task of the model and model characteristics.}
\label{tab:sota}
\end{table}

Table \ref{tab:sota} provides a summary of the relevant state-of-the-art approaches, highlighting their main properties and settings. Our examination reveals that existing research has predominantly concentrated on creating online classification models, which can operate with both stationary and drifting data streams. However, there has been comparatively limited effort devoted to online active linear regression or dedicated to constructing online regression models in general.

We believe that there are several promising directions for future research in this field. First, we recommend further investigation into online active learning strategies specifically designed for regression models. Given the limited work in this area, there is a need for more advanced methods that can be applied to nonlinear models, beyond linear models or linear bandits. For example, there has been a recent spark of interest toward the use of Bayesian optimization for active learning in nonlinear regression problems \citep{Mohamadi2020,Riis2022}. Additionally, model-agnostic methods that can be applied to a variety of regression models could be valuable as they would provide a more general solution to the problem. Second, we believe that there is potential for research into single-pass online sampling strategies for dynamic data streams. Ensemble models and batch-based approaches have been the dominant methods in online classification, but some of their assumptions or requirements may not hold in many real-world applications. For instance, in some applications, data may arrive in a continuous stream, and it may not be possible to divide it into batches due to time or memory constraints. In such cases, single-pass online sampling strategies that do not require the use or update of multiple models would be more practical. Moreover, it could be beneficial to develop online active learning strategies that are able to tackle all the types of distribution shifts introduced in Section \ref{subsec:drifting}. Finally, the combination of reinforcement learning and active learning in pool-based scenarios is an area of ongoing research. We believe that the study of online reinforcement learning to optimize sampling strategies could provide valuable insights into how to best perform active learning in dynamic environments.

\section{Conclusion} \label{sec:conclusion}
The field of online active learning with data streams is a rapidly evolving and highly relevant area of research in machine learning. The ability to effectively learn from data streams in real-time is becoming increasingly important, as the amount of data generated by modern applications continues to grow at an exponential rate. However, obtaining annotated data to train complex prediction and decision-making models presents a major roadblock. This hinders the proper integration of artificial intelligence models with real-world applications such as healthcare, autonomous driving and industrial production. Our survey provides a comprehensive overview of the current state of the art in this field and highlights the challenges and opportunities that researchers face when developing methods for online active learning. We reviewed a wide range of strategies for selecting the most informative data points in online active learning, including methods based on uncertainty sampling, diversity sampling, query by committee, and reinforcement learning, among others. Our analysis has shown that these strategies have been applied in a variety of contexts, including online classification, online regression, and online semi-supervised learning. We hope that this survey will inspire further research in the field of online active learning with data streams and encourage the development of new and advanced methods for handling this type of data. In particular, we believe that there is significant potential for the development of model-agnostic and single-pass online active learning strategies that can be applied in practical settings.

\section*{Acknowledgments}
The authors gratefully acknowledge the support of the DTU Strategic Alliances Fund, which made this research possible. We would also like to extend our sincere thanks to John S{\o}lve Tyssedal for his invaluable help and support throughout the project.









\typeout{}
\bibliography{sn-bibliography}

\end{document}